\documentclass[lettersize,journal]{IEEEtran}

\usepackage{cite}
\usepackage{graphicx}
\usepackage{amsfonts}
\usepackage{amsmath}
\usepackage{amssymb}
\usepackage{pifont}
\usepackage{bm,upgreek}
\usepackage{mathrsfs}
\usepackage{url}
\usepackage[linesnumbered, ruled]{algorithm2e}
\SetKwRepeat{Do}{do}{while}

\usepackage{pbox}

\usepackage[svgnames]{xcolor}
\usepackage{xstring}
\newcommand{\corr}{\textcolor{Black}}

\usepackage{array}
\usepackage{tabularray}
\usepackage{multirow}

\begin{document}

\title{Non-parametric regression for robot learning on manifolds}

\author{
    \IEEEauthorblockN{L{\'o}pez-Custodio, P.C.$^{*}$\thanks{$^{*}$Corresponding author.}\IEEEauthorrefmark{1}, Bharath, K.\IEEEauthorrefmark{2}, Kucukyilmaz, A.\IEEEauthorrefmark{1}, Preston, S.P.\IEEEauthorrefmark{2}}
    \\\
    \IEEEauthorblockA{\IEEEauthorrefmark{1}School of Computer Science, University of Nottingham}
    \\\
    \IEEEauthorblockA{\IEEEauthorrefmark{2}School of Mathematical Sciences, University of Nottingham
    \\\{pablo.lopez-custodio, karthik.bharath, ayse.kucukyilmaz, simon.preston\}@nottingham.ac.uk}
}

\maketitle

\begin{abstract}
    Many of the tools available for robot learning were designed for Euclidean data. However, many applications in robotics involve manifold-valued data. A common example is orientation; this can be represented as a 3-by-3 rotation matrix or a quaternion, the spaces of which are non-Euclidean manifolds. In robot learning, manifold-valued data are often handled by relating the manifold to a suitable Euclidean space, either by embedding the manifold or by projecting the data onto one or several tangent spaces. These approaches can result in poor predictive accuracy, and convoluted algorithms. In this paper, we propose an ``intrinsic'' approach to regression that works directly within the manifold. It involves taking a suitable probability distribution on the manifold, letting its parameter be a function of a predictor variable, such as time, then estimating that function non-parametrically via a ``local likelihood'' method that incorporates a kernel. We name the method kernelised likelihood estimation. The approach is conceptually simple, and generally applicable to different manifolds. We implement it with three different types of manifold-valued data that commonly appear in robotics applications. The results of these experiments show better predictive accuracy than projection-based algorithms. 
\end{abstract}


\section{Introduction}

In recent years, learning from demonstration (LfD) has
become an essential approach in robotics research. LfD has been particularly useful for imitating complex human motion which is difficult to mimic by trajectory planning algorithms. Instead, in LfD, the user teaches the robot to follow a trajectory by means of a series of demonstrations. 

In this context, Dynamic Movement Primitives (DMP) \cite{dmp} were introduced to learn a linear attractor system that follows a demonstration. However, in many applications, it is also desired to learn a probability distribution from a series of demonstrations. Encoding how ``important'' a point is, due to the variability of the demonstrations around it, is crucial in optimal control \cite{optimal_control}. This sense of variability can also be exploited to regulate the stiffness of the robot when it interacts with humans \cite{promp_stiff}, to avoid new obstacles \cite{gmm_obstacles}, or to detect intentions in shared control \cite{ayse_gere_intent}. Hence, probabilistic approaches have been proposed including Gaussian Mixture Regression (GMR) \cite{gmr,calinon_gmr}, Probabilistic Movement Primitives (ProMP) \cite{promp}, and Kernalised Movement Primitives (KMP) \cite{kmp}.

All these LfD methods rely on the multivariate normal distribution as the probabilistic model, and thus only work with Euclidean data. This is not a problem when working in joint space. However, in many applications, working in task space is more convenient. Working in task space aids awareness of the environment, facilitating obstacle avoidance and adaptation to changes. Working in task space is also convenient when learning human skills as it is possible to track objects or parts of the human body. Then these trajectories are transferable to different robots, whilst in joint-space learning, they are exclusive to the robot used for the demonstrations. 

However, the problem of not being able to represent orientations in Euclidean space poses a big challenge since the above-mentioned LfD methods are no longer suitable for the training data. A tempting fix is to work with Euler angles. However, since these are three directions, rather than three real numbers, the search space is $\mathbb{T}^3$, which is still non-Euclidean. Even worse, the gimbal-lock singularity means an infinity of triads can represent the same orientation which is a major problem for a probabilistic treatment.

\begin{figure}[ht]
    \centering
    \includegraphics[width=0.47\textwidth]{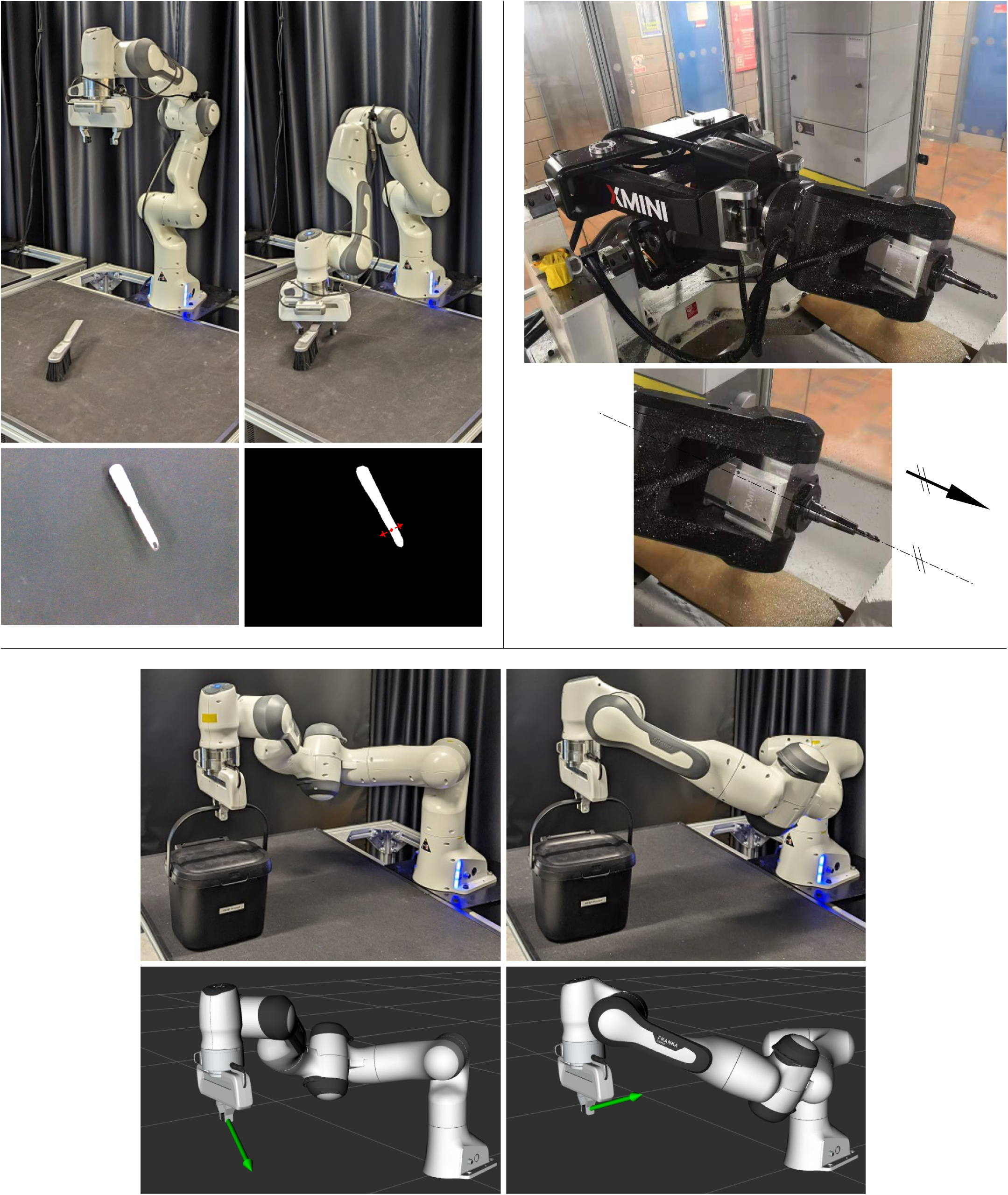}
    \caption{Examples of manifold-valued data in robotics. TOP-LEFT: A top-table grasp is an element of $\mathbb{R}^3\times\mathcal{S}^1$ which in this case parametrises the Schoenflies group of displacements, X(4). TOP-RIGHT: The orientation of the cutting tool of the 5-DOF Exechon XMini robot \cite{pc_exechon} is a direction in $\mathcal{S}^2$. BOTTOM: The 7-DOF Franka Panda robot lifts a box in two different configurations but the same pose of end-effector. The direction of maximum force achievable is shown for each configuration. This direction is found by computation of the manipulability ellipsoid, an element of the manifold of SPD matrices, $\mathrm{Sym}^{+}(6)$}
    \label{fig:intro}
\end{figure}

Unfortunately, orientations in task space are not the only type of non-Euclidean data encountered in robotics applications. Any 
data assuming values in a set where Euclidean geometry of vector spaces is not available is said to be non-Euclidean, 
 and in robotic applications the set can usually be modelled as a smooth manifold. Other examples of manifold-valued data in robotics include: the direction of a table-top grasp (Fig. \ref{fig:intro} top-left), the direction of a cutting tool as end-effector (Fig. \ref{fig:intro} top-right), the line defining a pointing action, the stiffness matrix of the robot, and the force and velocity manipulability ellipsoids (Fig. \ref{fig:intro} bottom). 

A common way to tackle the problem of working with manifold-valued data is through an embedding. Take for example data $\{\mathbf{x}_i\}_{i=1}^N$, where $\mathbf{x}_i$ are unit quaternions, so, per notation of Sec. ~\ref{sec:directions}, $\mathbf{x}_i \in \mathcal{S}^3/\mathbb{Z}_2$. If a ``mean orientation'' is of interest, the data is treated as Euclidean and the familiar arithmetic mean $N^{-1} \sum_i \mathbf{x}_i$ is evaluated. This results in a value in $\mathbb{R}^4$ that is not a unit quaternion. This can be remedied by renormalising, since
$\sum_i \mathbf{x}_i / \| \sum_i \mathbf{x}_i \|$ is a ``mean" that is indeed in the manifold. This is a simple example of what is sometimes called an \textit{extrinsic}, or \textit{embedding}, approach that embeds the manifold into a higher-dimensional Euclidean space, exploits the linear structure of that embedding space, then employs a projection mapping back into the manifold \cite{promp_stiff}. Extrinsic approaches such as this can work well, especially when the goal is to compute a point-wise quantity such as a mean, but for general manifolds there is not always a natural choice of embedding, nor way to characterise dispersion, and in some settings---such as when the data are highly dispersed on the manifold---estimators can behave erratically.

A different approach to analyse manifold-valued data is to employ \emph{tangent space} projections. This involves picking a point on the manifold, known as the base, and using the inverse exponential map to project the data from the manifold to the tangent plane of the base, which is a Euclidean space where standard tools can then be used. \corr{An advantage of this approach is that the toolbox available is very large, but a major disadvantage is that a tangent space projection may be inadequate in preserving the structure of the data.} Adequacy of the approach depends on two inter-related considerations: dispersion in the data around the base, and curvature of the manifold. Linearising the data through tangent space projections works well when dispersion is small and curvature of the manifold does not appreciably distort distances of points far from the base; when dispersion is large, the approach is suitable only when curvature of the manifold is non-positive everywhere. Large dispersion will result in distorted estimation of the dispersion structure of the projected data points on the tangent space of the base. 

Positively curved manifolds cannot be linearised through a single tangent space, and projections using the inverse exponential map are only locally well-defined. An important positively curved manifold that arises in robotics is the quotient manifold $\mathcal{S}^3/\mathbb{Z}_2$ of orientations used to represent three-dimensional rotations as quaternions with antipodal symmetry, where $\mathbf{x} \in \mathcal{S}^3$ and $-\mathbf{x} \in \mathcal{S}^3$ represent the same rotation. 

Relatedly, the Riemannian Gaussian distribution \cite{pennec,riemannian_gaussians} is commonly used in probabilistic modelling of manifold-valued data. The distribution is the image under the exponential map of a Gaussian distribution defined on the tangent space of a chosen base point, and operationally amounts to fitting a multivariate normal distribution to data projected onto the tangent space of the  base point. Again, curvature of the manifold plays an important role in its definition: for positively curved manifolds, support of the Gaussian distribution on the tangent space is restricted to a suitable subset, while no restrictions are required for non-positively curved manifolds. These distributions have been profitably used for manifold-valued data in robotics \cite{caldwell_projection,wang2022, calinon_gmr_manifolds,gmm_orientations, jaquier_spd, luis_manipulability,synergies,body_manifold}.

The above-mentioned issues for tangent space projections are exacerbated for data from trajectories on manifolds, since trajectories typically span larger subsets of the manifold, especially when considering one single tangent space \cite{caldwell_projection,wang2022}. In these cases, curvature effects become more pronounced, and data analysis based on tangent space projections can lead to unreliable conclusions. Notwithstanding these issues, adaptation of trajectory learning methods to manifold-valued data has been done relying on tangent-space projections\footnote{The adaptation of DMP to orientations \cite{orientational_dmp_1,orientational_dmp_2,orientational_dmp_3} does not require tangent-space projections since it does not provide a probabilistic model.}. These include the adaptations of ProMP \cite{orientational_promp} and KMP \cite{caldwell_projection} to orientational data, and GMR to general Riemannian manifolds \cite{calinon_gmr_manifolds,jaquier_spd}. \corr{The latter employs multiple tangent planes to tackle data covering large parts of the manifold. Hence, when using projections, researchers highly recommend against the use of a single tangent plane for the analysis of any manifold-valued data \cite{jaquier2024unraveling}}.

\begin{figure}[ht]
    \centering
    \includegraphics[width=0.40\textwidth]{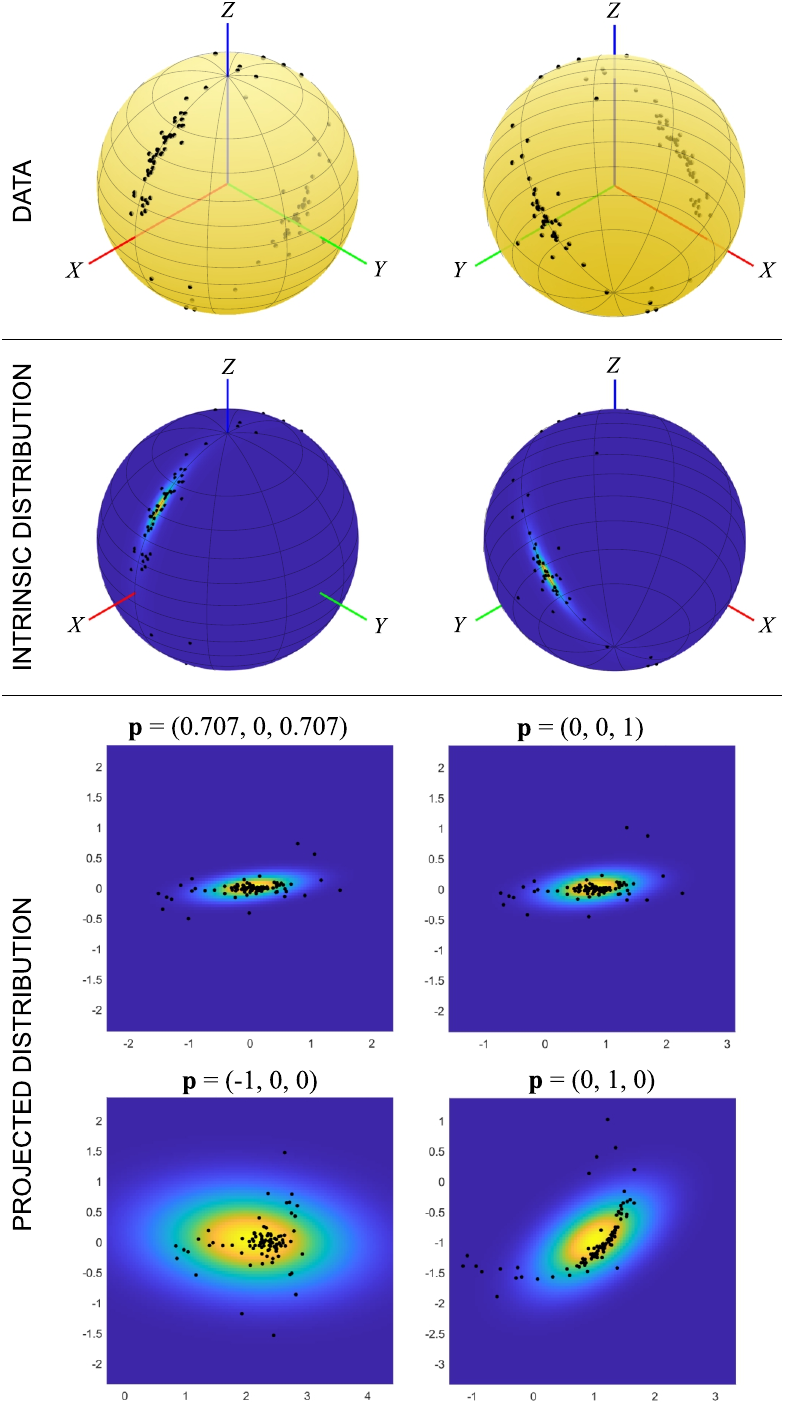}
    \caption{\corr{Visual comparion between an intrinsic distribution fitted to axial data in $\mathcal{S}^2/\mathbb{Z}_2$, and Gaussians fitted to projections of the data in different tangent spaces}}
    \label{fig:intr_vs_extrinsic}
\end{figure}

\corr{
In contrast to working in tangent spaces, {\it intrinsic} distributions model the data directly within the manifold. An example comparing an intrinsic distribution against a Gaussian projected in different tangent planes is shown in Fig. \ref{fig:intr_vs_extrinsic}. The data are axes in $\mathcal{S}^2/\mathbb{Z}_2$, i.e. $\mathbf{x}\in\mathcal{S}^2$ and $-\mathbf{x}$ are identified such that they represent the same element $\mathcal{S}^2/\mathbb{Z}_2$, in a similar way that the space of quaternions work. The data shown in the top row of Fig. \ref{fig:intr_vs_extrinsic} has mean $\mathbf{\upmu}=(\sqrt{2}/2,0,\sqrt{2}/2)$ and thus the data is concentrated around $\mathbf{\upmu}$ and $-\mathbf{\upmu}$. The first challenge for fitting a model in a tangent plane is to correct the sign of the data so that all of it can be projected. This is a straightforward if a good guess of $\mathbf{\upmu}$ is available, but otherwise the sign correction may lead to incorrect results. Considering this correction has been done correctly and the data is now concentrated only around $\mathbf{\upmu}$. The bottom of Fig. \ref{fig:intr_vs_extrinsic} shows Gaussians fitted at four different tangent planes with base points $\mathbf{p}\in\mathcal{S}^2$. The shape of these distributions is considerably dependant of $\mathbf{p}$. In addition, mapping of the means of these Gaussian back on to $\mathcal{S}^2$ results in different points on the sphere. In contrast, the second row of Fig. \ref{fig:intr_vs_extrinsic} shows an intrinsic distribution that was fitted to the data. This distribution is called Angular Central Gaussian (ACG) \cite{tyler_acg} and is explained in detailed in Sec. \ref{sec:acg}. This distribution is particularly well-suited for axial data, since no sign correction is required and, since no choice of tangent plane is needed, its MLE does not depend on the selection of any parameter and thus is always the same for a given data set. The two views of the distribution in the second row of Fig. \ref{fig:intr_vs_extrinsic} show its antipodal symmetry.}

Due to the problems arising from global linearisation of data via projections, in this paper we present an alternative method for robot learning that works directly on manifolds, i.e. it is intrinsic. Our approach exploits the idea of non-parametric regression, which remains largely unexplored in on-manifold robot learning research. In addition, we build upon probability distribution models that are rarely or never used in the robotics field. For example, we make use of the Angular Central Gaussian (ACG) distribution for axial data as a less computationally expensive alternative to the Bingham distribution \cite{bingham,mardia_jupp}, whose computational complexity makes it unappealing for robotics applications, although it has been used in the pose estimation problem \cite{bingham_vision} and in reinforcement learning \cite{bingham_rl}. Our approach is based on the idea of \emph{local likelihood} \cite{kernelised_likelihood,local_likelihood}, 
where regression is done by incorporating kernel weights into the terms of a log-likelihood expression. By this means, we are able to incorporate probability distributions tailored to the data structure in a regression model that relates a predictor variable to a manifold-valued response. We call our approach \emph{kernelised likelihood estimation}.

The advantages of our approach are the following:
\begin{itemize}
    \item The algorithm is \corr{conceptually} simple. The steps can be summarised in a maximization of a weighted sum of terms. This contrasts with other methods whose complexity stems from the adaptation of Euclidean-designed algorithms to manifold-valued data.
    \item As it will be shown in the experiments Section, by working intrinsically on the manifold, the estimation is more precise compared to methods that work on the tangent bundle. 
    \item The intrinsic nature of the estimation method ensures that it can be used for data on both positively and nonpositively curved manifolds.
    \item Both trajectory learning and regression can be carried out with the same algorithm.
    \item No joint distributions between predictor (input) and response (output) are defined. The shape of the covariance of these joint distributions is affected if either predictor or response are re-scaled, for example, due to a change of units. 
    \item The predictor is not limited to scalars, it can also be manifold-valued.
    \item In the case of trajectory learning, adaptation to via-points (both at the end and in the middle of the trajectory) is simple.  
    \item \corr{The model is quite robust to having a small number of training trajectories. This contrasts with other methods, like ProMP, in which a number of demonstrations smaller than the number of parameters in the model produce ill-conditioned covariance matrices. }
    \item A generative model for manifold-valued trajectories based on the learnt model is presented. This method reuses the same idea of kernelised likelihood.
    \item Unlike parametric models in which the model complexity is constrained by the parametrisation, the approach we propose incorporates non-parametric estimation, allowing the model to get more flexible as the amount of training data grows.
\end{itemize}

Model flexibility comes at the cost of new predictions having to depend directly on the training data (rather than some low-dimensional parameter vector); this is a potential disadvantage, though one shared by any non-parametric method. \corr{A comparison between different methods for probabilistic trajectory learning on manifolds is shown in Table \ref{tab:comparison}.}

\corr{There has been much recent interest and progress in probabilistic modeling of manifold-value data. Some examples not directly applicable to the problem of trajectory learning include approaches based on normalising flows (NF) \cite{NF}, diffusion models (DM) \cite{DF}, and energy-based (EB) \cite{EB} models. Variational formulations of these models towards approximating a target distribution typically mean that parameters of the approximating class of distributions (parameters of diffeomorphisms generating flows for NF, coefficients of SDEs for DM, and parameters of the energy function in EB) are not interpretable as characteristics of the trajectory distribution (e.g. average/mean trajectory, covariance and cross-covariance in a sample of trajectories). Another recent approach is the use of Riemannian Mat\'ern kernels \cite{Matern,Matern_jaquier}. However, those consider the very different problem in which the manifold is the domain and the response is scalar, whereas we in this paper we consider that response is manifold-valued. \cite{mallasto} consider wrapped Gaussian processes on manifolds, suited to manifold-valued response data, but in a way that requires having the mean trajectory computed as a pre-step separate from their approach, in contrast to our approach in which computing the mean trajectory is a central goal.
}

\begin{table*}[ht]
\caption{\corr{Comparison of probabilistic methods for trajectory learning on manifolds: Riemannian GMR \cite{calinon_gmr_manifolds}, Riemannian KMP \cite{caldwell_projection}, Orientation ProMP \cite{orientational_promp} and our method. }}
\label{tab:comparison}
\centering
\begin{tabular}{m{3cm} m{2cm} m{2cm} m{2cm} m{2cm}}
    \hline
    {} & \parbox{2cm}{\centering Does not rely on projection} & \parbox{2cm}{\centering Supports different manifolds} & \parbox{2cm}{\centering On-manifold predictor} & \parbox{2cm}{\centering Adaptation to via-points} 
    \\ 
    \hline
    Riemannian GMR & \parbox{2cm}{\centering \ding{55}} & \parbox{2cm}{\centering \checkmark} & \parbox{2cm}{\centering \checkmark} & \parbox{2cm}{\centering \ding{55}} 
    \\
    Riemannian KMP & \parbox{2cm}{\centering \ding{55}} & \parbox{2cm}{\centering \ding{55}} & \parbox{2cm}{\centering \ding{55}} & \parbox{2cm}{\centering \checkmark} 
    \\
    Orientation ProMP & \parbox{2cm}{\centering \ding{55}} & \parbox{2cm}{\centering \ding{55}} & \parbox{2cm}{\centering \ding{55}} & \parbox{2cm}{\centering \checkmark} 
    \\
    KLE (ours) & \parbox{2cm}{\centering \checkmark} & \parbox{2cm}{\centering \checkmark} & \parbox{2cm}{\centering \checkmark} & \parbox{2cm}{\centering \checkmark} 
    \\
    \hline
\end{tabular}
\end{table*}

In this paper, it is assumed that a task-space controller is available. Therefore, we only focus on the learning and generation of trajectories but not on the design of controllers.

This paper is organised as follows: Sec. \ref{sec:manifold_data} provides an overview of common manifold-valued data in robotics. Sec. \ref{sec:problem} defines the problem statement to be tackled in this paper. Sec. \ref{sec:nw} introduces the method we propose to solve this problem. In this method, a suitable probability distribution model is needed, hence, Sec. \ref{sec:models} presents several previously introduced models for different types of manifold-valued data. The proposed method is then tested in a series of experiments presented in Sec. \ref{sec:experiments}. Finally, conclusions are drawn in Sec. \ref{sec:conclusions}.

A video of the experiments carried out with the Franka Emika Panda arm is provided in the supplementary material accompanying this paper. The related source code will be made public upon acceptance of the paper. 


\section{Common examples of manifold-valued data in robotics}\label{sec:manifold_data}

Manifold-valued data appears in a multitude of applications in robotics. A list of these manifolds is summarised by \cite{calinon_manifolds_list}. Although the method presented in this paper works for any of those manifolds, we focus on the ones introduced in this Section, as these are some of the most commonly encountered manifolds in robotics applications.

\subsection{Directions, $\mathcal{S}^{d-1}$}\label{sec:directions}

Directions lie on the unit sphere $\mathcal{S}^{d-1}:=\{\mathbf{x}\,|\,\mathbf{x}^{\top}\mathbf{x}=1,\, \mathbf{x}\in\mathbb{R}^{d}\}$. An example of directional data is the orientation of a rotating cutting tool (Fig. \ref{fig:intro}, top-right), which is an element of $\mathcal{S}^{2}$. Similarly, the orientation of a table-top grasp is an angle in $\mathcal{S}^{1}$ (Fig. \ref{fig:intro}, top-left), while the whole grasp, including position, is in $\mathbb{R}^3\times\mathcal{S}^{1}$. Since the angle is always measured about an axis whose direction is constant, grasps belong to the 4-dimensional group of {\it Schoenflies} displacements, X(4) \cite{herveclass}. Other examples of data belonging to the X(4) group include the end-effector poses of the Delta and the SCARA robots.

\subsection{Axes, $\mathcal{S}^{d}/\mathbb{Z}_2$}\label{sec:axes}

In the manifold of axes, $\mathcal{S}^{d-1}/\mathbb{Z}_2$, $\mathbf{x}\in\mathcal{S}^{d-1}$ and its antipodal direction $-\mathbf{x}$ represent the same element. There are plenty of applications involving axial data, but an additional reason that $\mathcal{S}^{d-1}/\mathbb{Z}_2$ is so important is that it is isomorphic to the special orthogonal group, SO(3), defined in the following Section. Therefore, methods devised for $\mathcal{S}^{3}/\mathbb{Z}_2$ can be used directly for applications involving orientational data.

\subsection{Orientations, the special orthogonal group, $\mathrm{SO(3)}$; and poses, the special Euclidean group, $\mathrm{SE(3)}$} \label{sec:SO3:SE3}

The special orthogonal group of dimension 3, SO(3), is defined as the set of orthogonal matrices of determinant 1. These correspond to rotation matrices in 3-dimensional space. Hence, the orientation of any rigid body can be represented by an element of SO(3).

Consider the map $\mathbf{R}\,:\,\mathcal{S}^{3}\rightarrow\mathrm{SO}(3)$ defined as:
\begin{equation}\label{eq:quat2mat}    
\mathbf{R}(\mathbf{x}):=\mathbf{I}_3+2x_4\,\mathrm{skew}(x_1,x_2,x_3)+2\,\mathrm{skew}(x_1,x_2,x_3)^2
\end{equation}
\noindent where $\mathbf{x}:=(x_1,x_2,x_3,x_4)^{\top}\in\mathcal{S}^{3}$, and 
\[
    \mathrm{skew}(x_1,x_2,x_3):=\left(\begin{array}{ccc}0& -x_3 & x_2 \\
         x_3& 0 & -x_1 \\
         -x_2& x_1 & 0\\
    \end{array}\right).
\]

$\mathbf{R}(\cdot)$ is surjective, however $\mathbf{R}(\mathbf{x})=\mathbf{R}(-\mathbf{x})$, $\forall \mathbf{x}\in\mathcal{S}^{3}$. Due to this antipodal redundancy, SO(3) is isomorphic to $\mathcal{S}^{3}/\mathbb{Z}_2\cong P\mathbb{R}^3$. In kinematics, (\ref{eq:quat2mat}) provides a map between rotation matrices and unit quaternions.

The Special Euclidean group in 3-dimensional space, SE(3), is the set of rigid-body transformations. It is a Lie group, hence, it allows for a Lie algebra, se(3). SE(3) can be identified with the semidirect product $\mathrm{SE}(3)=\mathrm{SO}(3)\rtimes\mathbb{R}^3$.

The pose (position and orientation) of the end-effector of a robot with respect to a fixed coordinate system is a rigid body transformation and thus an element of SE(3). Therefore, for a task-space trajectory, the data are $\{(\mathbf{X}_i,t_i)\}_i^{N}$, where $\mathbf{X}_i\in\mathrm{SE}(3)$ is a 4-by-4 transformation matrix, a dual quaternion, or an element of any other parametrisation of SE(3).

For many applications, it is reasonable to work with position and orientation independently. Therefore, the task-space trajectory $\{(\mathbf{X}_i,t_i)\}_i^{N}$ can be represented as $\{(\mathbf{x}_i,\mathbf{p}_i,t_i)\}_i^{N}$, where $\mathbf{x}_i\in\mathcal{S}^{3}/\mathbb{Z}_2$ is an orientation, and $\mathbf{p}_i\in\mathbb{R}^3$, a position. 

\subsection{Symmetric positive definite matrices, $\mathrm{Sym}^{+}(d)$}\label{sec:spd}

The set of symmetric positive definite (SPD) matrices is defined as $\mathrm{Sym}^{+}(d):=\{\mathbf{M}\in\mathrm{Sym}(d)\,|\,\mathbf{x}^{\top}\mathbf{M}\mathbf{x}>0\,\forall \mathbf{x}\in\mathbb{R}^d\}$. 

Although data in $\mathrm{Sym}^{+}(d)$ appear in several robotics applications like stiffness and distance covariance matrices, we now explain one of the most important applications of $\mathrm{Sym}^{+}(d)$ in robot arms: the velocity and force manipulability ellipsoids.

Let $\mathbf{q}\in\mathbb{T}^r$ be the vector of joint angles of a robot arm with $r$ degrees of freedom (DOF). If the end-effector can move in a $d$-dimensional submanifold of SE(3), then its velocity $\mathbf{v}\in\mathbb{R}^d$ is related to the joint velocities by $\mathbf{v}=\mathbf{J}(\mathbf{q})\dot{\mathbf{q}}$, while the output wrench $\mathbf{f}\in\mathbb{R}^d$ is related to the joint torques $\bm{\uptau}\in\mathbb{R}^r$ by $\bm{\uptau}=\mathbf{J}^{\top}(\mathbf{q})\,\mathbf{f}$. The matrix $\mathbf{J}(\mathbf{q})\in\mathbb{R}^{d\times r}$ is the {\it Jacobian matrix} of the robot at configuration $\mathbf{q}$. 

It is possible to use $\mathbf{J}(\mathbf{q})$ to evaluate the performance of the robot at configuration $\mathbf{q}$ \cite{bruno}. One of such performance indexes evaluates the velocities of the end-effector that can be generated by the set of joint velocities of magnitude one, $\dot{\mathbf{q}}^{\top}\dot{\mathbf{q}}=1$ at $\mathbf{q}$. If $r>d$, replacing the least-squares solution to $\mathbf{v}=\mathbf{J}(\mathbf{q})\dot{\mathbf{q}}$, $\dot{\mathbf{q}} = \mathbf{J}^{\top}(\mathbf{q})(\mathbf{J}(\mathbf{q})\mathbf{J}^{\top}(\mathbf{q}))^{-1}\mathbf{v}$, in $\dot{\mathbf{q}}^{\top}\dot{\mathbf{q}}=1$ gives:
\begin{equation}\label{eq:manipul_ellipse}
    \mathbf{v}^{\top}\left(\mathbf{J}(\mathbf{q})\mathbf{J}^{\top}(\mathbf{q})\right)^{-1}\mathbf{v} = 1.
\end{equation}
Define $\mathbf{M}(\mathbf{q}):=\mathbf{J}(\mathbf{q})\mathbf{J}^{\top}(\mathbf{q})\in\mathrm{Sym}^{+}(d)$, and let $\{\lambda_i\}_{i=1}^{d}$ be its eigenvalues in descending order with corresponding eigenvectors $\{\mathbf{b}_i\}_{i=1}^{d}\subset\mathcal{S}^{d-1}$. Then (\ref{eq:manipul_ellipse}) defines an ellipsoid whose axes are parallel to $\{\mathbf{b}_i\}_{i=1}^{d}$ and have lengths $\{1/\sqrt{\lambda_i}\}_{i=1}^{d}$. Hence, $\mathbf{v}^{\top}\left(\mathbf{M}(\mathbf{q})\right)^{-1}\mathbf{v} = 1$ is known as the {\it velocity manipulability ellipsoid} at $\mathbf{q}$. Equivalently, $\mathbf{f}^{\top}\mathbf{M}(\mathbf{q})\mathbf{f} = 1$ provides a {\it force manipulability ellipsoid}.


\section{Probability distributions on relevant manifolds}\label{sec:models}

\corr{In this Section, we introduce probability distributions relevant to the manifolds introduced in Sec. \ref{sec:manifold_data}. For each distribution, we present the probability density function and the ``standard'' MLE solution given independent identically distributed data. These expressions are needed for the proposed method presented in Sec. \ref{sec:nw}.}


\subsection{The elliptically symmetric angular Gaussian distribution on $\mathcal{S}^{d-1}$}\label{sec:esag}

Given $\mathbf{y}\sim \mathcal{N}(\bm{\upmu},\mathbf{V})$, $\mathbf{y}\in\mathbb{R}^{d}$, with constraints $\mathbf{V}\bm{\upmu}=\bm{\upmu}$ and $|\mathbf{V}| = 1$, the directions $\mathbf{x}:=\mathbf{y}/\|\mathbf{y}\|\in\mathcal{S}^{d-1}$ have an {\it elliptically symmetric angular Gaussian} (ESAG) distribution with density \cite{esag}:
\begin{equation}\label{eq:esag}
\begin{split}
    f_{\mathrm{ESAG}}(\mathbf{x};\,\bm{\upmu},\,\mathbf{V}):=&\frac{C_d}{(\mathbf{x}^{\top}\mathbf{V}^{-1}\mathbf{x})^{d/2}}
    \mathrm{exp}\Bigg[\frac{1}{2}\Bigg\{\frac{(\mathbf{x}^{\top}\bm{\upmu})^2}{\mathbf{x}^{\top}\mathbf{V}^{-1}\mathbf{x}}
    \\&
    -\bm{\upmu}^{\top}\bm{\upmu}\Bigg\}\Bigg]
    M_{d-1}\left(\frac{\mathbf{x}^{\top}\bm{\upmu}}{(\mathbf{x}^{\top}\mathbf{V}^{-1}\mathbf{x})^{1/2}}\right),
\end{split}
\end{equation}

\noindent where $C_d:=1/(2\pi)^{(d-1)/2}$ and, for $d=2$ and 3, respectively:
\begin{eqnarray*}
    M_1(\alpha)&:=&\alpha\Phi(\alpha)+\phi(\alpha),
    \\
    M_2(\alpha)&:=&(1+\alpha^2)\Phi(\alpha)+\alpha\phi(\alpha),
\end{eqnarray*}
\noindent where $\phi(\cdot)$ and $\Phi(\cdot)$ are the functions for standard normal probability density and cumulative density, respectively. See \cite{esag} for $M_{d-1}$ with $d>3$.

Let $0\leq\rho_1,\leq\ldots\leq\rho_d=1$ be the eigenvalues of $\mathbf{V}$ with corresponding eigenvectors $\bm{\upxi}_1,\ldots,\bm{\upxi}_d$, where $\bm{\upxi}_d= \bm{\upmu}/\|\bm{\upmu}\|$. Then the mean of the distribution is $\bm{\upxi}_d$ while $\|\bm{\upmu}\|$ controls the concentration. The rest of the eigenvectors represent the axes of symmetry of the elliptical contours. 

For $d=3$, the ESAG distribution can be reparametrised to avoid the use of the constrained parameter $\mathbf{V}$. From these constraints, define $\rho$ so that $\rho_1=\rho$, $\rho_2=1/\rho$ and $\rho_3=1$. Define a new orthogonal frame $\{\tilde{\bm{\upxi}}_1,\tilde{\bm{\upxi}}_2,\tilde{\bm{\upxi}}_3\}$, where:
\begin{eqnarray*}
    \tilde{\bm{\upxi}}_3&:=&\bm{\upmu}/\|\bm{\upmu}\|,
    \\
    \tilde{\bm{\upxi}}_2&:=&(0,0,1)\times\bm{\upmu}/\|(0,0,1)\times\bm{\upmu}\|,
    \\
    \tilde{\bm{\upxi}}_1&:=&\tilde{\bm{\upxi}}_2\times\tilde{\bm{\upxi}}_3,
\end{eqnarray*}

\noindent where `$\times$' denotes cross product. Hence, the original eigenvectors can be obtained by rotating this new frame by an angle $\psi\in(0,\pi]$ about $\bm{\upmu}$, i.e. $\bm{\upxi}_i=\mathrm{Rot}(\psi,\bm{\upmu})\tilde{\bm{\upxi}}_i$, $i=1,2$. We then define $\bm{\upgamma}:=(\gamma_1,\gamma_2)$ by:
\[
    \gamma_1:=[(\rho^{-1}-\rho)/2]\cos 2\psi,\;\;\gamma_2:=[(\rho^{-1}-\rho)/2]\sin 2\psi.
\]
Consequently, the ESAG distribution for $d=3$ can be parameterised by $\bm{\upmu} \in \mathbb{R}^3$ and $\bm{\upgamma} \in \mathbb{R}^2$. For a data set $\{\mathbf{x}_i\}_{i=1}^N$, $\mathbf{x}_i\in\mathcal{S}^2$, assumed to come from ESAG, there is no closed-form solution for the ESAG MLE, hence numerical optimisation is needed to solve
\begin{equation}\label{eq:mle_esag}
    (\hat{\bm{\upmu}}, \hat{\bm{\upgamma}}) = \arg \max_{(\bm{\upmu}, \bm{\upgamma})} \sum_{i=1}^N\log f_{\mathrm{ESAG}}(\mathbf{x}_i;\, \bm{\upmu},\, \bm{\upgamma}).
\end{equation}


\subsection{The angular central Gaussian distribution on $\mathcal{S}^{d-1} /\mathbb Z_2$}\label{sec:acg}

Given $\mathbf{y}\sim \mathcal{N}(\mathbf{0},\mathbf{\Lambda})$, $\mathbf{y}\in\mathbb{R}^{d}$, the directions $\mathbf{x}:=\mathbf{y}/\|\mathbf{y}\|$ have an {\it angular central Gaussian distribution} (ACG) with density \cite{tyler_acg}:
\begin{equation}\label{eq:acg_def}
    f_\text{ACG}(\mathbf{x};\, \mathbf{\Lambda}):= \frac{\Gamma\left(\frac{d}{2}\right)}{2\sqrt{\pi^d\left|\mathbf{\Lambda}\right|}}\left(\mathbf{x}^{\top}\mathbf{\Lambda}^{-1}\mathbf{x}\right)^{-\frac{d}{2}},\;\; \mathbf{x}\in\mathcal{S}^{d-1} /\mathbb Z_2.
\end{equation}
The ACG is antipodally symmetric, i.e. $f_\text{ACG}(\mathbf{x};\, \mathbf{\Lambda})= f_\text{ACG}(-\mathbf{x};\, \mathbf{\Lambda})$, which makes it a candidate model for axial data. This distribution is uniquely defined up to scalar multiplication of the parameter $\mathbf{\Lambda}\in\mathrm{Sym}^{+}(d)$, i.e. $f_\text{ACG}(\mathbf{x};\,\mathbf{\Lambda}) = f_\text{ACG}(\mathbf{x};\,c\mathbf{\Lambda})$, $c\in\mathbb{R}^+$, an indeterminacy that can be resolved by imposing that $\mathrm{tr}(\mathbf{\Lambda}) = d$.

Consider the spectral decomposition $\mathbf{\Lambda} = \mathbf{B}\mathbf{A}\mathbf{B}^{\top}$, $\mathbf{B}:=\left[\mathbf{b}_{(1)},\ldots,\mathbf{b}_{(d)}\right]\in\mathrm{SO}(d)$, $\mathbf{A}:=\mathrm{diag}(a_1,\ldots,a_d)$, with eigenvalues $a_1,\ldots,a_d\in\mathbb{R}^{+}$ in descending order, and corresponding eigenvectors $\mathbf{b}_{(1)},\ldots,\mathbf{b}_{(d)}\in\mathcal{S}^{d-1}$. Then the mean of the ACG distribution is given by $\mathbf{b}_{(1)}$, and $a_1,\ldots,a_d$ represent the concentration in the corresponding principal directions. Due to the indeterminancy of $\mathbf{\Lambda}$, the eigenvalues are usually normalised so that $\mathrm{tr}(\mathbf{\Lambda})=a_1+\cdots + a_d=d$.

For a sample of axial data $\{\mathbf{x}_i\}_{i=1}^{N}$, $\mathbf{x}_i\in\mathcal{S}^{d-1}/\mathbb{Z}_2$, the MLE of the parameter $\mathbf{\Lambda}$ is defined implicitly as the solution of \cite{tyler_acg}:
\begin{equation}\label{eq:acg_mle_eq}
    \hat{\mathbf{\Lambda}} = \frac{d}{N}\sum_{i=1}^{N} \frac{\mathbf{x}_i\mathbf{x}_i^{\top}}{\mathbf{x}_i^{\top}\hat{\mathbf{\Lambda}}^{-1}\mathbf{x}_i}.
\end{equation}
In \cite{tyler_acg} it is proved that the iterative scheme,
\begin{equation}\label{eq:acg_mle}
    {\mathbf{\Lambda}}_{k+1} = d\displaystyle\sum_{i=1}^{N}\left\{ \dfrac{\left(\mathbf{x}_i^{\top}{\mathbf{\Lambda}}_k^{-1}\mathbf{x}_i\right)^{-1}}{\sum\limits_{i=1}^{N}\left(\mathbf{x}_i^{\top}{\mathbf{\Lambda}}_k^{-1}\mathbf{x}_i\right)^{-1}}\mathbf{x}_i\mathbf{x}_i^{\top}\right\},
\end{equation}
starting from $\mathbf{\Lambda}_{0} = \bm{I}$ converges as $k$ increases to the solution $\hat{\mathbf{\Lambda}}$ of (\ref{eq:acg_mle_eq}).
A practical algorithm is to continue iterating until $d_{\mathrm{SPD}}(\hat{\mathbf{\Lambda}}_{k+1},\hat{\mathbf{\Lambda}}_{k})\leq \epsilon$, where $d_{\mathrm{SPD}}(,)$ is a metric in $\mathrm{Sym}^{+}(d)$ and $\epsilon$ is a desired small value.


\subsection{The SPD lognormal distribution on $\mathrm{Sym}^{+}(d)$}\label{sec:spd_ln}

The cone $\mathrm{Sym}^{+}(d)$ of $d \times d$ symmetric positive definite matrices is a manifold whose tangent space at any point can be identified with $\mathrm{Sym}(d)$, the vector space of $d$-by-$d$ symmetric matrices. 
When equipped with the log-Euclidean metric, $\mathrm{Sym}^{+}(d)$ has non-positive curvature and is \emph{geodesically complete} \cite{schwartzman}. This means that geodesics between any two points are unique, and any point in $\mathrm{Sym}^{+}(d)$ can be uniquely mapped to a point in $\mathrm{Sym}(d)$. The nonpositive curvature ensures that, unlike the cases of the other manifolds mentioned in Sec. \ref{sec:manifold_data}, it is possible to define a probability distribution with full support on the tangent space of an arbitrary point without needing to ``wrap'' its image onto the manifold under the exponential map.

Consider the tangent space at the identity, then a point $\mathbf{X}\in\mathrm{Sym}^{+}(d)$ with spectral decomposition $\mathbf{X}=\mathbf{B}\mathbf{A}_1\mathbf{B}^{\top}$ is mapped to $\mathrm{Sym}(d)$ by the logarithmic map:
\begin{equation}
    \mathrm{log}(\mathbf{X}) = \mathbf{B}\,\mathrm{log}(\mathbf{A}_1)\mathbf{B}^{\top}\in\mathrm{Sym}(d),
\end{equation}
\noindent where log($\cdot$) acts element-wise. The corresponding exponential map for $\mathbf{Y}=\mathbf{B}\mathbf{A}_2\mathbf{B}^{\top}\in\mathrm{Sym}(d)$ is given by:
\begin{equation}
    \mathrm{exp}(\mathbf{Y}) = \mathbf{B}\,\mathrm{exp}(\mathbf{A}_2)\mathbf{B}^{\top}\in\mathrm{Sym}^{+}(d),
\end{equation}
\noindent where exp($\cdot$) acts element-wise.

Since $\mathrm{Sym}(d)$ is a $\left(\frac{1}{2}d(d+1)\right)$-dimensional vector space, it is possible to define the {\it symmetric-matrix normal distribution} as follows:
\begin{equation}
    \mathbf{Y}\sim\mathcal{N}_{\mathrm{Sym}}(\mathbf{M},\mathbf{\Sigma})\;\Leftrightarrow\;
    \mathrm{vecd}(\mathbf{Y})\sim\mathcal{N}(\mathrm{vecd}(\mathbf{M}),\mathbf{\Sigma}),
\end{equation}
\noindent where $\mathbf{Y},\mathbf{M}\in\mathrm{Sym}(d)$, $\mathbf{\Sigma}\in\mathrm{Sym}^{+}\left(\frac{1}{2}d(d+1)\right)$, and 
\[
    \mathrm{vecd}(\mathbf{Y}):=\left(\mathrm{diag}(\mathbf{Y})^{\top},\,\sqrt{2}\,\mathrm{offdiag}(\mathbf{Y})^{\top}\right)^{\top}\in\mathbb{R}^{\frac{1}{2}d(d+1)}.
\]

We now define the {\it lognormal distribution} in $\mathrm{Sym}^{+}(d)$ as follows \cite{schwartzman}:
\begin{equation}\label{eq:lognormal}
    \mathbf{X}\,\sim\,LN(\mathbf{M},\mathbf{\Sigma}) \;\Leftrightarrow\; \mathrm{log}(\mathbf{X})\,\sim\,\mathcal{N}_{\mathrm{Sym}}(\mathrm{log}(\mathbf{M}),\mathbf{\Sigma}),
\end{equation}
\noindent where $\mathbf{X},\mathbf{M}\in\mathrm{Sym}^{+}(d)$, and $\mathbf{\Sigma}\in\mathrm{Sym}^{+}\left(\frac{1}{2}d(d+1)\right)$. Hence, $\mathbf{X}\,\sim\,LN(\mathbf{M},\mathbf{\Sigma})$ has the following density:
\begin{equation}
\begin{split}
    f_{\mathrm{LN}}(\mathbf{X};\,\mathbf{M},\mathbf{\Sigma}):=\frac{J(\mathbf{X})}{N_{\mathrm{LN}}}\mathrm{exp}\Bigg\{-\frac{1}{2}\mathrm{vecd}\big[\mathrm{log}(\mathbf{X})
    &\\[1ex]
    -\mathrm{log}(\mathbf{M})\big]^{\top}
    \mathbf{\Sigma}^{-1}\mathrm{vecd}\big[\mathrm{log}(\mathbf{X})-\mathrm{log}(\mathbf{M})\big]\Bigg\},
    &
\end{split}
\end{equation}
\noindent where the normalizing term is given by $N_{\mathrm{LN}}=(2\pi)^{\frac{1}{4}d(d+1)}|\mathbf{\Sigma}|^{\frac{1}{2}}$, and $J(\mathbf{X})$ is the Jacobian in Eq. (14) of \cite{schwartzman}.

Due to the simplicity of the definition of the lognormal distribution in (\ref{eq:lognormal}), the maximum likelihood estimate of the parameters is straightforward. Given a sample of SPD data $\{\mathbf{X}_i\}_{i=1}^{n}$, $\mathbf{X}_i\in\mathrm{Sym}^{+}(d)$, this estimation is given by:
\begin{eqnarray}\label{eq:spd_ln_mle_m}
\hat{\mathbf{M}}&=&\mathrm{exp}\left(\frac{1}{n}\sum_{i=1}^{n}\mathrm{log}\left(\mathbf{X}_i\right)\right),
\\ \label{eq:spd_ln_mle_cov}
\hat{\mathbf{\Sigma}}&=&\frac{1}{n}\sum_{i=1}^{n} \mathrm{vecd}\left(\mathrm{log}(\mathbf{X}_i)-\mathrm{log}(\hat{\mathbf{M}})\right)
\nonumber\\ &{}&\times 
\mathrm{vecd}\left(\mathrm{log}(\mathbf{X}_i)-\mathrm{log}(\hat{\mathbf{M}})\right)^{\top}.
\end{eqnarray}


\section{Problem statement}\label{sec:problem}

In the probabilistic trajectory-learning problem, it is desired to predict a variable (response) for a new input (predictor) given a training dataset representing a series of human demonstrations. It is also required to know the variability of the demonstrations for any value of the predictor. We consider the cases in which either the response or both the response and the predictor lie on a manifold. 

In the case of scalar predictor and response in a manifold $\mathcal{M}$, the training data $\mathscr{D}=\{\{(\mathbf{x}_{ij},t_{ij})\}_{j=1}^{M_i}\}_{i=1}^{N}$, $t_{ij}\in[0,1]$, $\mathbf{x}_{ij}\in\mathcal{M}$, is a set of $N$ trajectories recorded from human demonstrations, each trajectory $i$ contains $M_i$ data points. Let $P_{\mathcal{M}}(\theta)$ be a probability distribution on \corr{the manifold that the data belongs to}, $\mathcal{M}$, with parameter(s) $\theta$, so that the mean and dispersion structure are directly or indirectly included in $\theta$. Then, \corr{given $\mathscr{D}$}, it is desired to estimate $\hat{\theta}(t)$, for any $t\in[0,1]$. So, we can obtain a trajectory of $L$ points, $\{\hat{\theta}_k,t_k\}_{k=1}^{L}$, $\hat{\theta}_k:=\hat{\theta}(t_k)$. 

It is also desired to solve the problem of simple regression. In this case, given a new value of the predictor, the value of the response is estimated based on a training dataset. Hence, the training data set is $\mathscr{D}=\{(\mathbf{x}_{i},\mathbf{t}_{i})\}_{i=1}^N$, $\mathbf{x}_{i}\in\mathcal{M}_O$, $\mathbf{x}_{i}\in\mathcal{M}_I$, where $\mathcal{M}_I$ and $\mathcal{M}_O$ are the manifolds of the predictor and response, respectively. The goal is now to estimate $\hat{\mathbf{x}}\in\mathcal{M}_O$ for a new $\mathbf{t}\in\mathcal{M}_I$.

\corr{Note that, in this paper the double subscript is used for the case of data sets consisting of multiple trajectories, e.g. $\{\{(\mathbf{x}_{ij},t_{ij})\}_{j=1}^{M_i}\}_{i=1}^{N}$, where $t_{ij}$ is the $j-$th time-step of the $i-$th trajectory. Otherwise, a single subscript is used for a single trajectory, e.g. $\{\mu_i,t_i\}_{i=1}^{L}$, where $t_{i}$ is the $i-$th time-step of the mean trajectory. In addition, a single subscript is also used if the data is a set of points rather than trajectories, e.g. $\{(\mathbf{x}_{i},\mathbf{t}_{i})\}_{i=1}^N$, where $\mathbf{t}_{i}$ is the value of the predictor of the $i-$th point.}


\section{Robot learning via kernelised likelihood estimation} \label{sec:nw}

The approach we develop in this paper extends upon the idea of Nadaraya--Watson kernel regression \cite{nadaraya,watson}, a non-parametric approach that estimates the response variable for a predictor query point as weighted combinations of the training response cases, with weights chosen according to proximity of predictor values. While Nadaraya--Watson regression is appropriate for Euclidean-valued response data, our approach is appropriate for manifold-valued response data and hence to the problems described in Sec. \ref{sec:problem}.  Furthermore, Nadaraya--Watson regression is usually geared towards computing a point estimate, such as a mean, for each value of the predictor, whereas here we are interested too in estimating the dispersion structure.

\subsection{Nadaraya-Watson regression in Euclidean space}

Consider training data $\mathscr{D} = \{(x_i,t_i)\}_{i=1}^N$, $x_i,t_i\in\mathbb{R}$. The Nadaraya--Watson estimator of the mean response as a function of the predictor is
\begin{equation}\label{eq:basic}
\hat{\mu}(t) = \sum_{i=1}^N\left[\dfrac{K_h(t-t_{i})}{\sum\limits_{i=1}^{N}K_h(t-t_{i})}x_{i}\right],
\end{equation}
where $K_h(t)$ is a non-negative-valued kernel function with bandwidth parameter $h>0$, a maximum at $t=0$, and $\int K_h(t) \mathrm{d} t$ equal to a constant independently of $h$. The value of $h$ sets a trade-off between fitting and smoothness. A commonly used kernel is defined as $K_h(t):=\mathrm{exp}(-t^2/2h^2)$.

Writing $(\ref{eq:basic})$ as
\begin{equation}\label{eq:basic:in:terms:of:W}
\hat{\mu}(t) = \sum_{i=1}^N W_{i}(t) \, x_{i},
\end{equation}
emphasises that the estimator is a simple weighted linear combination of the training response data, with the weight
\begin{equation}\label{eqn:weights:defn}
  W_{i}(t) := \dfrac{K_h(t-t_{i})}{\sum\limits_{i=1}^{N} K_h(t-t_{i})}.
\end{equation}

\subsection{Kernelised likelihood estimation}

The Nadaraya--Watson estimator (\ref{eq:basic:in:terms:of:W}) is appealing because it is intuitive and simple to compute. However, it does not generalise directly to manifold-valued response data, $\mathbf{x}_i \in \mathcal{M}$, because weighted combinations of $\mathbf{x}_i$ do not in general lie in $\mathcal{M}$. 
Instead, consider first a probability distribution $P_{\mathcal{M}}(\theta)$ defined on the manifold $\mathcal{M}$, with probability density function $f(\mathbf{x} ; \theta)$ and parameter(s) $\theta$, and write
\begin{equation}\label{eq:rv:X:on:M}
X \sim P_{\mathcal{M}}(\theta).
\end{equation}

Given a sample of data $\{\mathbf{x}_i\}_{i=1}^N$ assumed to be realisations of (\ref{eq:rv:X:on:M}) independent of each other and from $t_i$, parameter $\theta$ can be estimated from the data by maximum likelihood estimation (MLE). The MLE of $\theta$ is the maximiser of the log-likelihood:
\begin{equation}\label{eq:general_mle}
    \hat{\theta} = \arg \max_{\theta} \sum_{i=1}^N\log f(\mathbf{x}_i;\, \theta).
\end{equation}

In this paper, instead of the data being independent they are of the paired form $\{(\mathbf{x}_i, t_i)\}_{i=1}^N$, and the goal is to predict how the response variable $\mathbf{x}_i$ depends on the predictor $t_i$. \corr{To this end we consider models of the following form for $X$ given a value of the predictor $t$:}

\begin{equation}\label{eq:model:X:conditional:on:T}
(X \vert T = t) \sim P_{\mathcal{M}}(\theta(t)). 
\end{equation}

This is still somewhat general, because it remains to specify how $\theta(t)$ depends on $t$. We impose this dependence by prescribing
\begin{equation}\label{eq:weighted_mle}
    \hat{\theta}(t) = \arg \max_{\theta} \sum_{i=1}^{N}W_{i}(t)\log f(\mathbf{x}_{i};\, \theta),
\end{equation}
where $W_i(t)$ is as defined in (\ref{eqn:weights:defn}). \corr{Clearly, $\hat\theta(t)$ depends in general on the definition of the chosen kernel and its hyperparameters, as explained after Eq. (\ref{eq:basic})}. This is maximising a ``local likelihood" in the sense of \cite{local_likelihood}. Reflecting how the local likelihood incorporates kernel weights, we call our approach \emph{kernelised likelihood estimation} (KLE).

Once an estimate $\hat{\theta}(t)$ is computed from training data then, by (\ref{eq:model:X:conditional:on:T}), an estimate of $\mathbb{E}(X|T=t)$ is
\begin{equation} \label{eqn:mu_hat:as:fn:of:theta:hat}
    \hat{\bm{\upmu}}(t) = \mathbb{E}\big(P_{\mathcal{M}}\big(\hat{\theta}(t)\big)\big),
\end{equation}
where $\mathbb{E}$ denotes expectation. 
Thus, $\hat{\bm{\upmu}}(t)$ defines a smooth path through the training data and, in this sense, provides a solution to the on-manifold regression problem. 

For the problem of trajectories on manifolds with training data $\mathscr{D}=\{\{(\mathbf{x}_{ij},t_{ij})\}_{j=1}^{M_i}\}_{i=1}^{N}$, $t_{ij}\in[0,1]$, $\mathbf{x}_{ij}\in\mathcal{M}$, we simply rewrite Eq. (\ref{eq:weighted_mle}) as:
\begin{equation}\label{eq:weighted_mle_traj}
    \hat{\theta}(t) = \arg \max_{\theta} \sum_{i=1}^{N}\sum_{i=1}^{M_i}W_{ij}(t)\log f(\mathbf{x}_{ij};\, \theta),
\end{equation}
\noindent where
\begin{equation}\label{eq:kernel_weight_ij}
    W_{ij}(t):=\dfrac{K_h(t-t_{ij})}{\sum\limits_{i=1}^{N}\sum\limits_{j=1}^{M_i}K_h(t-t_{ij})}.
\end{equation}

To illustrate the application of the KLE method, consider the following scenarios with scalar, Euclidean and manifold-valued data:

{\bf Scalar response:} Consider the case of $x_i, t_i \in \mathbb{R}$ (that is, let $\mathcal{M}$ be $\mathbb{R}$), let $P_{\mathcal{M}}(\theta)$ with $\theta = \{\mu, \sigma^2\}$ be the normal distribution $\mathcal{N}(\mu, \sigma^2)$. Then, by Eq. (\ref{eq:weighted_mle}), $\hat{\theta}(t) = \{\hat{\mu}(t), \hat{\sigma}^2(t)\}$, where
\begin{equation}\label{eq:scalar}
\hat{\mu}(t) = \sum_{i=1}^N W_{i}(t) x_{i}, \quad \hat{\sigma}^2(t) = \sum_{i=1}^N W_{i}(t) [x_{i} - \hat{\mu}(t)]^2.
\end{equation}
Thus in this case the estimator of the mean, $\hat{\mu}(t)$, is exactly the Nadaraya-Watson estimator (\ref{eq:basic:in:terms:of:W}). This approach further provides an estimate of how the variance, ${\sigma}^2$, varies with the predictor, $t$. 

{\bf Euclidean response:} Now consider $\mathscr{D}=\{\{(\mathbf{x}_{ij},t_{ij})\}_{j=1}^{M_i}\}_{i=1}^{N}$, $\mathbf{x}_{ij}\in\mathbb{R}^d$, $t_{ij}\in[0,1]$ (that is, let $\mathcal{M}$ be $\mathbb{R}^d$), and let $P_{\mathcal{M}}(\theta)$ with $\theta = \{\bm{\upmu}, {\bm{\Sigma}}\}$ be the multivariate normal distribution $\mathcal{N}(\bm{\upmu}, {\bm{\Sigma}})$. Then $\hat{\theta}(t) = \{\hat{\bm{\upmu}}(t), \hat{{\bm{\Sigma}}}(t)\}$, where
\begin{equation}\label{eq:nw_normal}
\begin{gathered}
    \hat{\bm{\upmu}}(t) = \sum_{i=1}^N\sum_{j=1}^{M_i}W_{ij}(t)\,\mathbf{x}_{ij},\\
    \hat{\bm{\Sigma}}(t) = \sum_{i=1}^N\sum_{j=1}^{M_i}\left[W_{ij}(t)\,\left(\mathbf{x}_{ij}-\hat{\bm{\upmu}}(t)\right)
    \left(\mathbf{x}_{ij}-\hat{\bm{\upmu}}(t)\right)^{\top}\right],
\end{gathered}
\end{equation}
where the weights $W_{ij}(t)$ Eq. (\ref{eq:kernel_weight_ij}). \corr{Hence in the special case with $\mathcal{M}$ being Euclidean and $P_{\mathcal{M}}$ being Gaussian, KLE produces very intuitive estimators that are simple weighted versions of the standard MLEs.} The proof of Eq. (\ref{eq:nw_normal}) is given in \ref{sec:appendix}. 

{\bf Manifold-valued predictor:} In the case of a manifold-valued predictor, $\mathbf{t}_{ij}\in\mathcal{M}_I$, Eq. (\ref{eq:weighted_mle}) holds with a kernel of the form $K_h(d_{\mathcal{M}_I}(\mathbf{t}, \mathbf{t}_{ij}))$, where $d_{\mathcal{M}_I}(\cdot,\cdot)$ is a metric on $\mathcal{M}_I$.

\corr{
{\bf Manifold-valued response:} In this case, $\mathscr{D}=\{\{(\mathbf{x}_{ij},t_{ij})\}_{j=1}^{M_i}\}_{i=1}^{N}$, $\mathbf{x}_{ij}\in\mathcal{M}$, $t_{ij}\in[0,1]$. A suitable model for the distribution in $\mathcal{M}$ is chosen. The probability distributions presented in Sec. \ref{sec:models} admit the following solutions for Eq. (\ref{eq:weighted_mle}):
}

\begin{itemize}
\item \corr{{\bf ESAG:} Given $\mathcal{M}=\mathcal{S}^{2}$ and based on (\ref{eq:mle_esag}), the following numerical optimisation must be solved for each $t$:
\begin{eqnarray*}
    &{}&(\hat{\bm{\upmu}}(t), \hat{\bm{\upgamma}}(t)) = 
    \\
    &{}&\arg \max_{(\bm{\upmu}, \bm{\upgamma})} \sum_{i=1}^N
    \sum_{j=1}^{M_i}
    W_{ij}(t)
    \log f_{\mathrm{ESAG}}
    (\mathbf{x}_{ij};\, \bm{\upmu},\, \bm{\upgamma}).
\end{eqnarray*}
}
\item \corr{{\bf ACG:} Given $\mathcal{M}=\mathcal{S}^{d-1}/\mathbb{Z}_2$, the solution to (\ref{eq:weighted_mle}) satisfies
\begin{equation}\label{eq:acg_kernel_lik}
    \hat{\mathbf{\Lambda}}(t) = d\sum_{i=1}^{N} 
    \sum_{i=j}^{M_i}
    W_{ij}(t)
    \frac{\mathbf{x}_{ij}\mathbf{x}_{ij}^{\top}}{\mathbf{x}_{ij}^{\top}\hat{\mathbf{\Lambda}}(t)^{-1}\mathbf{x}_{ij}}.
\end{equation}
Derivation of this kernelised analogue of (\ref{eq:acg_mle_eq}) is given in \ref{sec:appendix}. 
An iterative scheme for finding the solution
$\hat{\mathbf{\Lambda}}(t)$ of (\ref{eq:acg_kernel_lik}) at a new input $t\in[0,1]$ is
\begin{equation}\label{eq:nw_acg}
\begin{split}
    &{\bm{\Lambda}}_{k+1}(t) = 
    \\
    &d\sum_{i=1}^{N}\sum_{j=1}^{M_i}\left[ \dfrac{K_h(t-t_{ij})\left(\mathbf{x}_{ij}^{\top}{\bm{\Lambda}}^{-1}_k\mathbf{x}_{ij}\right)^{-1}\mathbf{x}_{ij}\mathbf{x}_{ij}^{\top}}{\sum\limits_{i=1}^{N}\sum\limits_{j=1}^{M_i}K_h(t-t_{ij})\left(\mathbf{x}_{ij}^{\top}{\bm{\Lambda}}^{-1}_k\mathbf{x}_{ij}\right)^{-1}}\right]
\end{split}
\end{equation}}

\item \corr{{\bf SPD Lognormal:} Given $\mathcal{M}=\mathrm{Sym}^{+}(d)$, the estimate of $\hat{\mathbf{M}}$ and $\hat{\mathbf{\Sigma}}$ for a new input $t\in[0,1]$ is obtained by inserting kernels in Eqs. (\ref{eq:spd_ln_mle_m}) and (\ref{eq:spd_ln_mle_cov}), respectively:
\begin{equation}\label{eq:nw_spd}
    \hat{\mathbf{M}}(t)=\mathrm{exp}\left(\sum_{i=1}^{N}\sum_{j=1}^{M_i} W_{ij}(t)\mathrm{log}(\mathbf{X}_{ij})\right),
\end{equation}
\begin{equation}
    \hat{\mathbf{\Sigma}}(t)=\sum_{i=1}^{N}\sum_{j=1}^{M_i} W_{ij}(t)\bm{\uptau}_{ij},
\end{equation}
\noindent where
\[
    \begin{split}
    \bm{\uptau}_{ij}&:=\mathrm{vecd}\left(\mathrm{log}(\mathbf{X}_{ij})-\mathrm{log}(\hat{\mathbf{M}})\right)
    \\&\qquad\times
    \mathrm{vecd}\left(\mathrm{log}(\mathbf{X}_{ij})-\mathrm{log}(\hat{\mathbf{M}})\right)^{\top}.
    \end{split}
\]
Derivation of Eq. (\ref{eq:nw_spd}) is given in \ref{sec:appendix}.}
\end{itemize}

\corr{
To illustrate the application of the KLE method to on-manifold trajectory learning, consider the pseudocode in Algorithm \ref{alg:acg_traj_learn}. In this case, $\mathscr{D}=\{\{(\mathbf{x}_{ij},t_{ij})\}_{j=1}^{M_i}\}_{i=1}^{N}$, $\mathbf{x}_{ij}\in\mathcal{S}^{d-1} /\mathbb Z_2$, $t_{ij}\in[0,1]$, that is, $\mathcal{M}$ is $\mathcal{S}^{d-1} /\mathbb Z_2$. To model the data, an ACG distribution (with parameter $\bm{\Lambda}$) is used. In Algorithm \ref{alg:acg_traj_learn}, the function \texttt{Ker\char`_MLE\char`_ACG} can be replaced by the corresponding solution to the KLE in Eq. (\ref{eq:weighted_mle}) for a probability distribution that is suitable for the data. It can be seen that Algorithm \ref{alg:acg_traj_learn} has a simpler structure than most of the other methods for on-manifold trajectory learning. 
}

\begin{algorithm}
    \DontPrintSemicolon
    \SetKwFunction{FMLE}{\texttt{Ker\char`_MLE\char`_ACG}}
    \SetKwProg{Fn}{Function}{:}{\KwRet $\hat{\bm{\Lambda}}_{\mathrm{new}}$}
    \Fn{\FMLE{$t$, $\mathscr{D}$, $h$, $\varepsilon$}}{
        $\hat{\bm{\Lambda}}_{\mathrm{old}} = \mathbf{I}_d$\;
        \Do{$E<\varepsilon$}{
            $\hat{\bm{\Lambda}}_{\mathrm{new}}$ $\leftarrow$ \corr{Solution of } Eq.(\ref{eq:nw_acg}) \tcp*[f]{input: $t,\mathscr{D}, h$}\;
            $E = d_{\mathrm{SPD}}(\hat{\bm{\Lambda}}_{\mathrm{old}},\hat{\bm{\Lambda}}_{\mathrm{new}})$\;
            $\hat{\bm{\Lambda}}_{\mathrm{old}} = \hat{\bm{\Lambda}}_{\mathrm{new}}$\;
        }
    }
    \;
    \SetKwFunction{FMain}{main}
    \SetKwProg{Fn}{Function}{:}{\KwRet $\{t_n,\hat{\bm{\Lambda}}_n\}_{n=1}^{T}$}
    \Fn{\FMain{$\mathscr{D}$, $T$, $h$, $\varepsilon$}}{
        \For{$n = 1$ \KwTo $T$}{
            $t_n = (n-1)/(T-1)$\;
            $\hat{\bm{\Lambda}}_n =\,$\texttt{Ker\char`_MLE\char`_ACG}$\,(t_n,\mathscr{D},h,\varepsilon)$\;
        }
    }
  \caption{Trajectory learning with ACG distribution}\label{alg:acg_traj_learn}
\end{algorithm}

\subsection{Trajectory adaptation to via-points, and blending of trajectories}\label{sec:adaptation}

Consider the simplest case of a scalar trajectories dataset $\mathscr{D}=\{\{(x_{ij},t_{ij})\}_{j=1}^{M_i}\}_{i=1}^N$, $x_{ij}\in\mathbb{R}$, $t_{ij}\in[0,1]$. Then $\hat{\mu}(t)$ and $\hat{\sigma}(t)$, $t\in[0,1]$, can be determined by Eq (\ref{eq:scalar}). It is desired to adapt this model so that the trajectories pass through $\mu^{*}$ at $t^{*}$ with an uncertainty controlled by $\sigma^{*}$. 

\corr{
To adapt this model, we synthetically generate a second dataset $\mathscr{E}=\{(y_{i},t^{*})\}_{i=1}^{S}$, where $y_{i}\sim\mathcal{N}(\mu^{*},\sigma^{*})$. Then, the adapted model is obtained as a mixture of $\mathscr{D}$ and $\mathscr{E}$. The kernels weighting $\mathscr{D}$ are now multiplied by an activation function $\alpha(t)$ that smoothly ignores the points in $\mathscr{D}$ around $t^{*}$ pulling the mean estimation towards $\mathscr{E}$. Hence, the adapted mean yields:
}

\begin{figure}[ht]
    \centering
    \includegraphics[width=0.48\textwidth]{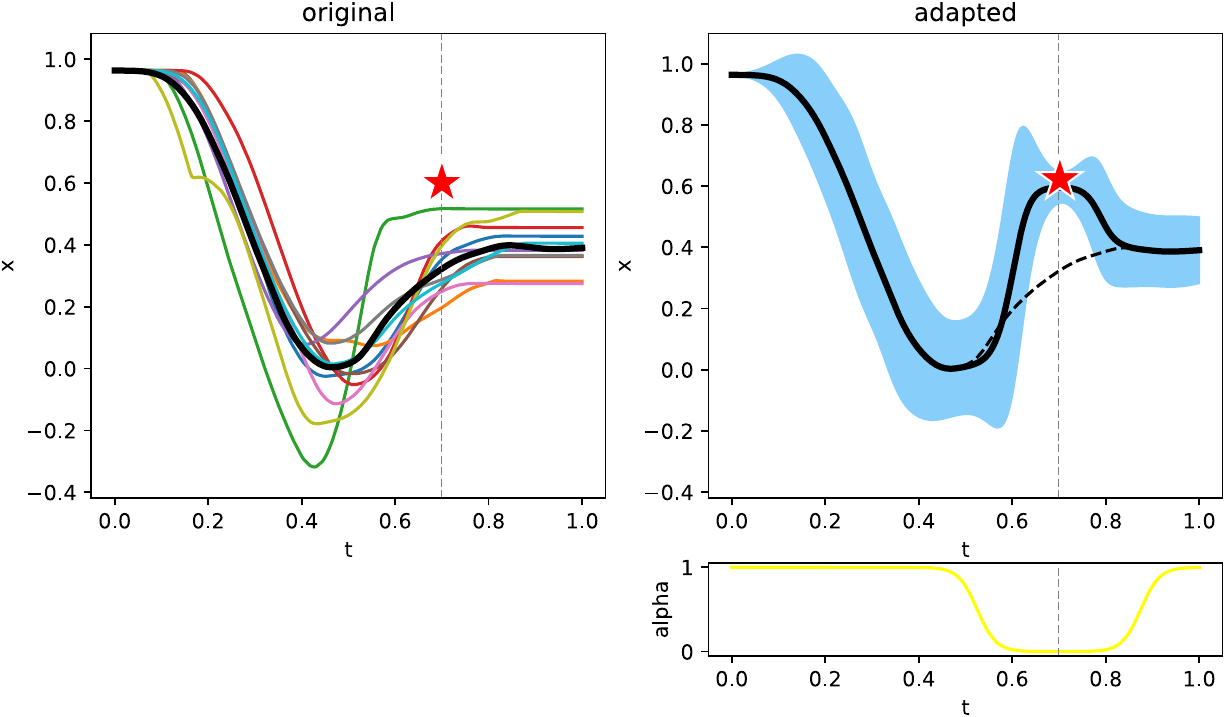}
    \caption{\corr{Adaptation of scalar trajectories to $\mu^{*}=0.6$ at $t^{*}=0.7$. LEFT: $\mathscr{D}$ consists of 700 samples taken from 10 demonstrated curves. The resulting $\hat{\mu}(t)$ is shown in thick black curve. $\mathscr{E}$ is $S=10$ samples generated with $\mu^{*}=0.6$ and $\sigma^{*}=0.01$. RIGHT: The result of adaptation with the blue area showing $\pm 1.5\hat{\sigma}(t)$. BOTTOM RIGHT: The activation function correspond to $\alpha(t)=1-0.5\tanh(25(t-t^{*}+0.175))+0.5\tanh(25(t-t^{*}-0.175))$}}
    \label{fig:conditioning}
\end{figure}

\corr{
\begin{eqnarray*}
    \hat{\mu}(t) &=& 
    \frac{1}{C(t)}
    \left[\sum\limits_{i=1}^N\sum\limits_{j=1}^{M_i}\alpha(t_{ij})K_h(t-t_{ij})x_{ij}
    \right.\\&{}&\left.
    +\sum\limits_{i=1}^{S} K_g(t-t^{*})y_{i} \right],
\end{eqnarray*}
\noindent where
\[
    C(t) = \sum\limits_{i=1}^N\sum\limits_{j=1}^{M_i}\alpha(t_{ij})K_h(t-t_{ij})
    +\sum\limits_{i=1}^{S}K_g(t-t^{*}),
\]
\noindent where $\alpha(t^{*})\approx0$ while $\alpha(t)\approx1$ for $t$ far away from $t^{*}$. The adapted $\hat{\sigma}(t)$ is obtained inserting the same activation functions. If adaptation to multiple points is required, the activation function has to be designed to deactivate the points in $\mathscr{D}$ around the new desired via-points. Fig. \ref{fig:conditioning} shows an example in which a series of scalar demonstrations are adapted to $\mu^{*}=0.6$ at $t^{*}=0.7$.
}

\corr{
The method requires the selection of $g$, $S$ and $\alpha(t)$. If the latter has, for example, the form $\tanh{(a(t-b))}$, $a$ and $b$ have to be chosen too. If the desired via-point is far from the training data, the selection of these parameters sets a trade-off between the smoothness of the transition to the via-point, and the inflation of the covariance during this transition. This is due to the way kernelised non-parametric methods carry out estimation. However, if the new via-point is close or inside the training data this inflation of covariance does not occur. It is well known that extrapolation of trajectory-learning models represents more challenges \cite{deep_adaptation}. The effects of manipulating these parameters in a scalar adaptation task are analysed in \ref{sec:appendix_adaptation}.
}

Since this modification to the KLE algorithm simply acts on the kernels \-- by weighting them with the activation functions \-- the same change is applied to the kernels in Eq (\ref{eq:weighted_mle}) in order to adapt trajectories on manifolds to new via-points.

The problem of trajectory blending can be solved by the same method by letting the activation functions act on the data points of the two trajectory sets to be blended. In this case, the activation functions are smoothed step functions.


\subsection{Trajectory generation}\label{sec:generation}

After $\hat{\theta}(t)$ has been computed for a trajectories dataset $\mathscr{D}=\{\{(\mathbf{x}_{ij},t_{ij})\}_{j=1}^{M_i}\}_{i=1}^{N}$, $\mathbf{x}_{ij}\in\mathcal{M}$, $t_{ij}\in[0,1]$, new trajectories can be generated around the mean by re-sampling and smoothing along $t\in[0,1]$. First, generate a set $\mathscr{D}_{r}=\{(\mathbf{x}_{r,i},t_{r,i})\}_{i=1}^{N_r}$, where $\mathbf{x}_{r,i}\sim P_{\mathcal{M}}(\hat{\theta}(t_{r,i}))$. Then, apply the KLE regression from Eq. (\ref{eq:weighted_mle}) as a smoother for $\mathscr{D}_{r}$. This results in the estimation of the parameter $\hat{\theta}_r(t)$ for $\mathscr{D}_{r}$. Then, the generated curve, $\hat{\bm{\upmu}}_r(t)$, is defined by Eq. (\ref{eqn:mu_hat:as:fn:of:theta:hat}), by taking
$\hat{\theta}(t)$ equal to $\hat{\theta}_r(t)$. 
Consider the special case when $\mathcal{M}$ is $\mathbb{R}^d$, $P_\mathcal{M}$ is the normal distribution, and $\hat{\theta}_r(t)$ in Eq. (\ref{eqn:mu_hat:as:fn:of:theta:hat}) is treated as random (depending on the sampling distribution of the training data) then $\hat{\bm{\upmu}}_r(t)$ is a Gaussian Process (GP). Hence the proposed approach to trajectory generation is a generalization to $\mathcal{M}$ of a GP model in $\mathbb{R}^d$; see \ref{sec:appendix} for further explanation. \corr{It has to be mentioned that if multiple trajectories are generated, their variability may be smaller in comparison to the training set. This arises because simulating a trajectory by simulating then smoothing multiple trajectories involves averaging that shrinks the variance compared with the simulated trajectories.}


\subsection{Autonomous systems}

For trajectories in task space, velocities can be learnt as well in order to generate time-independent trajectories. Hence, the training data for an autonomous system is $\mathscr{D}=\{\{(\mathbf{T}_{ij},\mathbf{V}_{ij},t_{ij})\}_{j=1}^{M_i}\}_{i=1}^{N}$, where $\mathbf{T}_{ij}\in\mathrm{SE}(3)$ is a pose and $\mathbf{V}_{ij}\in\mathrm{se}(3)$ a twist. The proposed on-manifold learning method allows to learn both the position and the orientation parts of $\mathbf{T}_{ij}$, and since se(3) is a vector space, $\mathbf{V}_{ij}$ is easy to learn with the same method using the multivariate normal distribution as model (Eq. (\ref{eq:nw_normal})). Finally, the generated trajectories have the form $\{(\mathbf{T}_{n},\mathbf{V}_{n})\}_{n=1}^{T}$. Since se(3) enjoys of Euclidean structure, the learning of $\mathbf{V}_{ij}$ is not taken into account in the rest of the paper.


\section{Experiments}\label{sec:experiments}

In this Section, four experiments are presented to test the KLE algorithm. The experiments involve data that comes from the different manifolds introduced in Sec. \ref{sec:manifold_data}. In the first two cases, our results are compared against those obtained with the PbDLib project \cite{calinon_github,calinon_tutorial} which works with GMR. \corr{In addition, a GMR model fitted in a single 
 tangent plane, as suggested in \cite{caldwell_projection}, is also compared in the first experiment.} In the experiments involving the Franka Emika Panda arm, the coordinate systems $O$ and $E$ are coincident with the frames \texttt{O} and \texttt{EE} in the LibFranka environment \cite{libfranka}. All values of distances are in meters, while angles are in radians. \corr{As per (\ref{eq:quat2mat}), our quaternion convention is $(x,y,z,w)\in\mathcal{S}^{3}/\mathbb{Z}_2$.}


\subsection{Trajectories on $\mathcal{S}^2$ with scalar predictor}\label{sec:exp_1}

Consider the manifold $\mathcal{S}^2$ which was introduced in Sec. \ref{sec:directions}. In this experiment, the KLE method is applied to the problem of trajectory learning on $\mathcal{S}^2$. Hence, it is desired to obtain a mean and a dispersion structure for new values of the scalar predictor.

Consider $\mathscr{D}=\{\{(\mathbf{x}_{ij},t_{j})\}_{j=1}^{M}\}_{i=1}^{N}$, $\mathbf{x}_{ij}\in\mathcal{S}^{2}$, $t_{j}\in[0,1]$. $\mathscr{D}$ consists of $N=10$ hand-written demonstrations of the letter B which are projected onto the unit sphere. Each trajectory $i=1,\ldots,10$ contains $M=100$ time-stamped points. These demonstrations are taken from the dataset available from the PbDLib package \cite{calinon_github}. 

\begin{figure}[ht]
    \centering
    \includegraphics[width=0.47\textwidth]{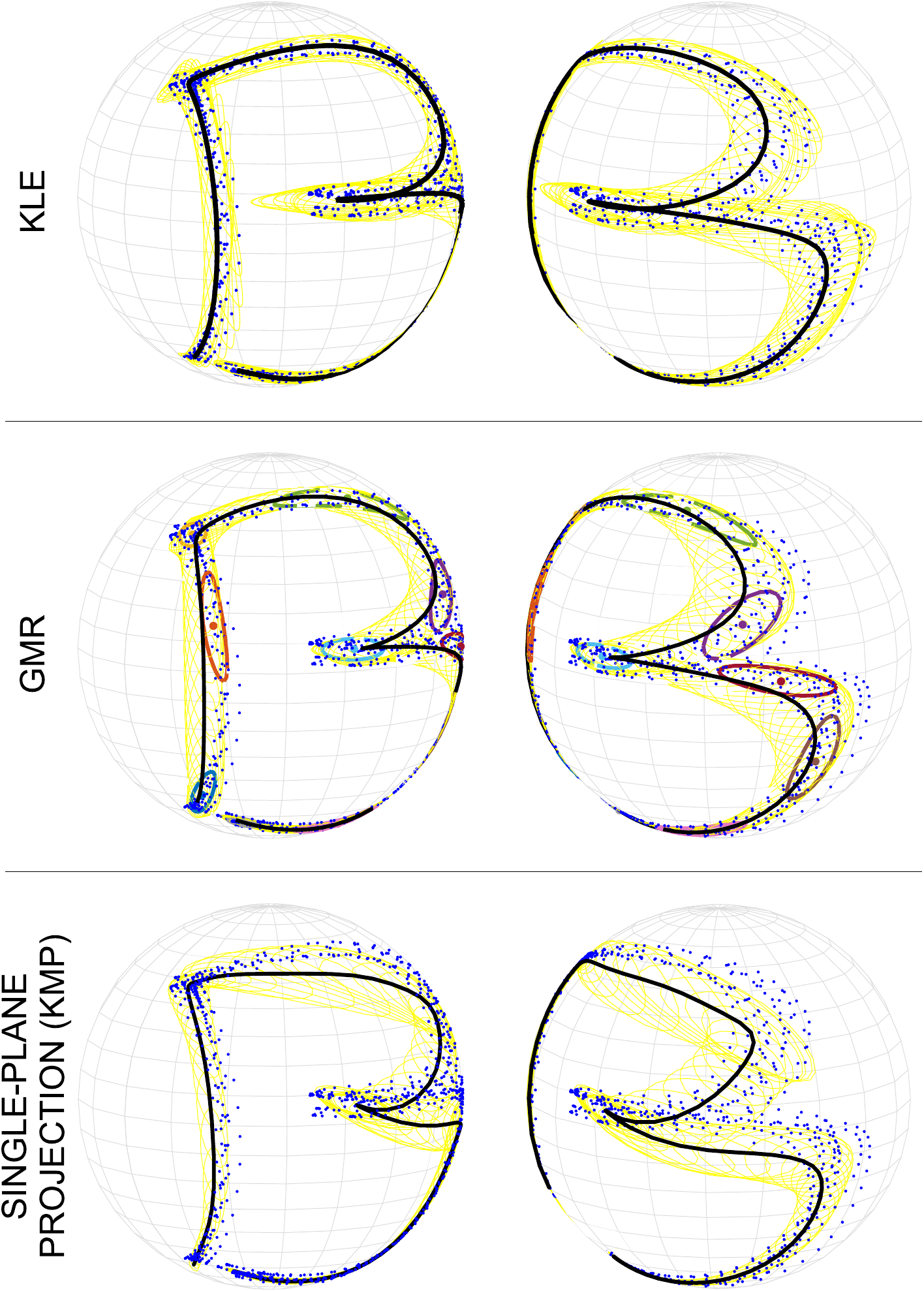}
    \caption{\corr{Regression for trajectories on $\mathcal{S}^2$. TOP: KLE method with ESAG model. MIDDLE: Riemannian GMR with 10 stages, the covariance for each stage is shown in coloured thicker curves. BOTTOM: KMP with 10 stages.}}
    \label{fig:ex_s2}
\end{figure}

The KLE algorithm is applied using the ESAG distribution on $\mathcal{S}^2$ as model. The parameters $\hat{\bm{\upmu}}(t)$ and $\hat{\bm{\upgamma}}(t)$ are estimated for 100 equidistant values of $t\in[0,1]$ using $h = 0.01$. The result is shown in the top row of Fig. \ref{fig:ex_s2}. \corr{As baselines for comparison, we consider the Riemannian GMR \cite{calinon_gmr_manifolds} and the Riemannian KMP \cite{caldwell_projection} approaches.} The code for Riemannian GMR, available from PbDLib \cite{calinon_github}, is run for the same dataset $\mathscr{D}$. The number of stages for the GMR method is fixed to 10. The results are shown in the second row of Fig. \ref{fig:ex_s2}. \corr{For the Riemannian KMP we used the code available in \cite{kmp_github}. This method first obtains a GMR model in a single-plane projection, then this is used as base curve for a kernelised regression model. Since this is an approximation of the base curve, we use the results of the GMR step as baseline. The same dataset is run with 10 stages for the GMM, the start of the first trajectory is used as base point for the tangent space as suggested in \cite{caldwell_projection}. The results are shown in the bottom row of Fig. \ref{fig:ex_s2}}.

\corr{An inspection of Fig. \ref{fig:ex_s2} shows that the mean (thick black curve) predicted by the KLE method with ESAG model passes through the data more {\it centrally} than that predicted by the Riemannian GMR and the Riemannian KMP methods. The shape and orientation of the probability level ellipses (yellow curves) predicted by KLE vary more in shape and orientation than those predicted by both baselines, revealing more flexibility in the model. To evaluate the performance of the mean estimation, we first consider the mean squared error (MSE) at each time step and obtain the total error along the duration of the trajectory:}
\corr{
\begin{equation}\label{eq:e1}
    e_1 := \sum_{j=1}^{M}\left\{\frac{1}{N}\sum_{i=1}^{N}\mathrm{acos}\big(\hat{\bm{\upmu}}(t_{j})^{\top}\mathbf{x}_{ij}\big)^2\right\}
\end{equation}
}
\corr{Secondly, we evaluate the distance of the estimated mean to the extrinsic mean of the data at each time step and take the total error along the trajectory:
\begin{equation}\label{eq:e2}
    e_2 := \sum_{j=1}^{M}\mathrm{acos}\big(\hat{\bm{\upmu}}(t_{j})^{\top}\hat{\bm{\upnu}}_{j}\big),\,\,\hat{\bm{\upnu}}_{j}:=\left.\sum_{i=1}^{N}\mathbf{x}_{ij}\middle/\left|\sum_{i=1}^{N}\mathbf{x}_{ij}\right.\right|
\end{equation}
Table \ref{tab:s2_results} shows the results of applying these two evaluation criteria.
}

\begin{center}
\begin{table}[h]
\caption{\corr{Performance comparison for KLE (our method), and GMR and KMP baselines. The second column shows the total MSE error computed using (\ref{eq:e1}), while the third column shows the error to the extrinsic mean, (\ref{eq:e2})}}
\label{tab:s2_results}
\centering
\begin{tabular}{ccc}
    \hline
    {} & MSE & Distance to extrinsic mean
    \\ 
    \hline
    KLE & 4.8668 & 1.7390
    \\
    GMR & 5.5534 & 7.6633
    \\
    KMP & 5.4985 & 6.4726
    \\
    \hline
    {} 
\end{tabular}
\end{table}
\end{center}

\subsection{Trajectories on $\mathrm{Sym}^{+}(2)$ with scalar predictor}\label{sec:exp_2}

In order to test learning of trajectories on the manifold $\mathrm{Sym}^{+}(d)$, which was introduced in Sec. \ref{sec:spd}, at a dimension that allows visualization of the data, we consider $d=2$. It is desired to obtain a mean and a dispersion structure for new values of the predictor. For this experiment we generate synthetic data, this allows us to have a ground truth against which the results can be compared.

We generate synthetically the set $\mathscr{D}=\{\{(\mathbf{x}_{ij},t_{j})\}_{j=1}^{M}\}_{i=1}^{N}$, $\mathbf{x}_{ij}\in\mathrm{Sym}^{+}(2)$, $t_{j}\in[0,1]$. For every $t_j$ we simulate the LN distribution, i.e. $\mathbf{x}_{ij}\sim LN(\mathbf{M}_j,\bm{\Sigma}_j)$, $j=1,\ldots,N$. By controlling the parameters $\mathbf{M}_j\in\mathrm{Sym}^{+}(2)$ and $\bm{\Sigma}_j\in\mathrm{Sym}^{+}(3)$ we have knowledge of the ground truth.

Consider the spectral decomposition $\mathbf{M}_j=\mathbf{B}(t_j)\,\mathrm{diag}(a_1(t_j),a_2(t_j))\,\mathbf{B}(t_j)^{\top}$, then $\mathbf{B}(t)$ and $(a_1(t),a_2(t))$ are trajectories in $\mathrm{SO}(2)$ and $\mathbb{R}^2$, respectively. We generate $\mathbf{B}(t)$ as planar rotation matrices parametrized by the angle of rotation $\beta(t)$. Then, $\beta(t)$, $a_1(t)$ and $a_2(t)$ require simple scalar interpolation.  

Similarly, the decomposition $\mathbf{\Sigma}_j=\mathbf{C}(t_j)\,\mathrm{diag}(b_1(t_j),b_2(t_j),b_3(t_j))\,\mathbf{C}(t_j)^{\top}$ requires the interpolation of scalars $b_1(t),b_2(t),b_3(t)\in\mathbb{R}$, while $\mathbf{C}(t)\in\mathrm{O}(3)$ can be generated by spherical linear interpolation (Slerp) between $\mathbf{C}(0)$ and $\mathbf{C}(1)$.

For this particular experiment, for each of $M=15$ equidistant $t_j$ we generated $N=30$ samples. The following scalar interpolations were considered:
$a_1(t) = (1+9t)^2$, $a_2(t) = (2-t)^2$, $\beta(t) = 6.6322t-6.2831t^2$, $b_1(t) = 0.2+1.8t$, $b_2(t)=0.15+0.85t$ and $b_3(t)=0.1$, where $\beta(t)$ was chosen so that $\beta(0)=0$, $\beta(0.5)=100\pi/180$ and $\beta(1)=20\pi/180$. The slerp interpolation for $\mathbf{C}(t)$ was done between $\mathbf{C}(0)=\mathbf{I}_3$ and $\mathbf{C}(1)=\mathbf{R}_Z(15\pi/180)\mathbf{R}_Y(5\pi/180)\mathbf{R}_X(10\pi/180)$.

\begin{figure}[ht]
    \centering
    \includegraphics[width=0.47\textwidth]{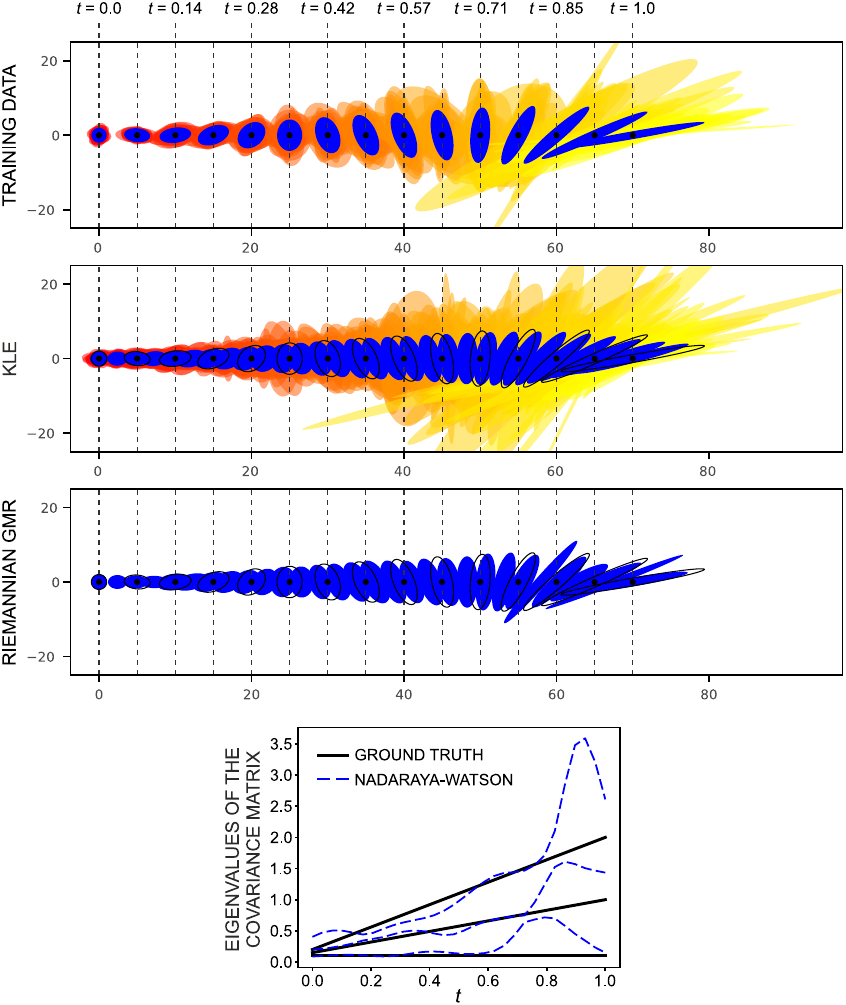}
    \caption{Learning of trajectories on $\mathrm{Sym}^{+}(2)$. FIRST PLOT: Training data $\mathscr{D}$ (orange to yellow) and original mean $\mathbf{M}_j$ (blue). SECOND PLOT: Using KLE: estimated mean $\hat{\mathbf{M}}(t)$ (blue), original mean $\mathbf{M}_j$ (black outline), and samples from $LN(\hat{\mathbf{M}}(t), \hat{\bm{\Sigma}}(t))$ (orange to yellow). THIRD PLOT: Using Riemannian GMR: estimated mean $\hat{\mathbf{M}}(t)$ (blue), and original mean $\mathbf{M}_j$ (black outline). BOTTOM: eigenvalues of the original covariance matrix $\bm{\Sigma}_j$ (solid black), and KLE-computed $\hat{\bm{\Sigma}}(t)$ (dashed blue)}
    \label{fig:ex_spd2}
\end{figure}

The first plot from top to bottom of Fig. \ref{fig:ex_spd2} shows $\mathscr{D}$. The samples are ellipses which are concentric for each value of $t_j$ and their colour progresses from orange at $t_0=0$ to yellow at $t_{15}=1$. The 15 means $\mathbf{M}_j$ are shown in blue ellipses. We ran the proposed KLE approach with the estimators in Eq. (\ref{eq:nw_spd}) using $h=0.06$. We computed 30 queries at equidistant values of $t\in[0,1]$. The results are shown in the second plot of Fig. \ref{fig:ex_spd2}. The estimated mean $\hat{\mathbf{M}}(t)$ is shown in blue ellipses, while the original means, $\mathbf{M}_j$, are superposed in black outline for comparison. For each request, 30 samples are drawn from $LN(\hat{\mathbf{M}}(t),\hat{\mathbf{\Sigma}}(t))$, their colour progresses from orange (at $t=0$) to yellow (at $t=1$). 

It can be seen that the estimated $\hat{\mathbf{M}}(t)$ follows the trend in both shape and orientation of original $\mathbf{M}_j$. The samples drawn from the estimated LN distributions also follow the pattern shown by the training dataset, with an expected slightly larger uncertainty. This uncertainty can be visualised in the bottom plot of Fig. (\ref{fig:ex_spd2}), where the eigenvalues of the estimated covariances $\hat{\bm{\Sigma}}(t)$ (dashed blue curves) are plotted against those of the original $\bm{\Sigma}_j$. It can be seen that the uncertainty introduced by the sampling from the original distributions, and the regression process perturb the eigenvalues. However, the trend of those still matches the original ones.

We ran the code for Riemannian GMR available from the PbDLib \cite{calinon_github, calinon_tutorial} for the same dataset $\mathscr{D}$ with 10 stages. The results are shown in the third plot of Fig. \ref{fig:ex_spd2}. The estimated mean $\hat{\mathbf{M}}(t)$ is shown in blue ellipses while the original means $\mathbf{M}_j$ are the superposed outlined ellipses. It can be seen that the mean follows the trend almost in a similar way to our approach. However, at the end of the trajectory, the mean is more distorted than the one estimated by KLE. Although the Riemannian GMR method uses the same LN distribution as model, we were unable to obtain comparable results for the covariance $\hat{\bm{\Sigma}}(t)$. Hence, its eigenvalues and the samples from the estimated LN distributions are not presented here.

\corr{
Table \ref{tab:spd2_results} shows a comparison between the results obtained using KLE and Riemannian GMR. Since for this experiment the ground truth is known, we compute accumulated error of the estimated mean with respect to that of the ground truth at each time step using the square log-Euclidean distance as metric:
\begin{equation}\label{eq:e_spd}
    e := \sum^M_{j=1}\mathrm{tr}\left[\left(\mathrm{log}(\hat{\mathbf{M}}(t_j))-\mathrm{log}(\mathbf{M}_j)\right)^2\right]
\end{equation}
}

\begin{center}
\begin{table}[h]
\caption{\corr{Performance comparison for KLE (our method) and GMR as baseline. The second column shows the error computed using (\ref{eq:e_spd}), i.e. total distance to the ground-truth mean.}}
\label{tab:spd2_results}
\centering
\begin{tabular}{cc}
    \hline
    {} & Distance to ground-truth mean 
    \\ 
    \hline
    KLE & 0.53822
    \\
    GMR & 1.08549
    \\
    \hline
    {} 
\end{tabular}
\end{table}
\end{center}

\subsection{Trajectories on $\mathrm{SE(3)}$ with scalar predictor}\label{sec:exp_3}

\begin{figure}[ht]
    \centering
    \includegraphics[width=0.47\textwidth]{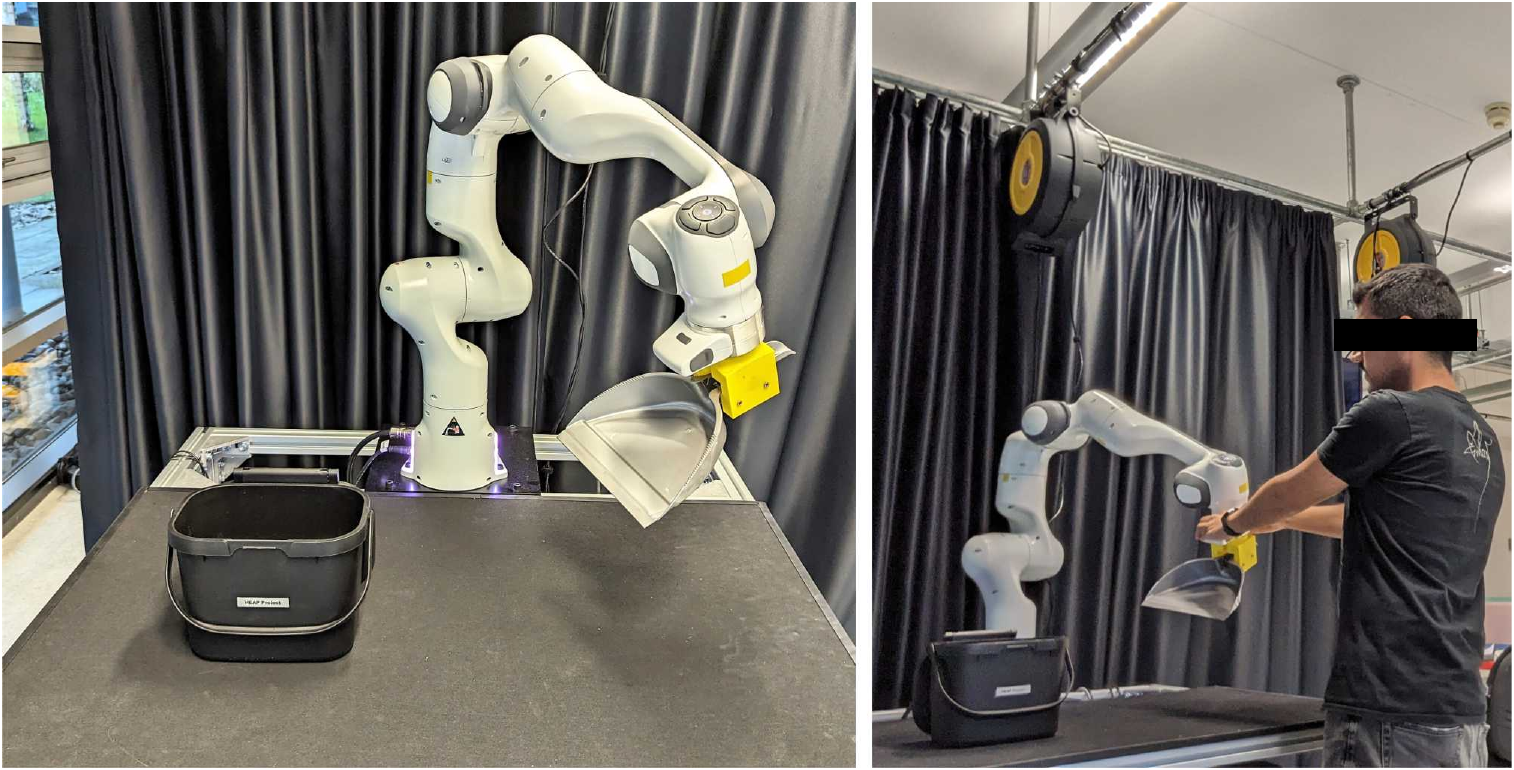}
    \caption{LEFT: Setup for experiment in Sec. \ref{sec:exp_3} -- a Franka Panda arm with a dustpan as end-effector, and a bin on the table. RIGHT: A participant carries out a kinesthetic demonstration.}
    \label{fig:dustpan_setup}
\end{figure}

In this experiment, we teach a Franka Emika Panda arm with a dustpan attached to its end-effector to drop rubbish into the bin, see Fig. \ref{fig:dustpan_setup} (left). The demonstrated trajectories involve large changes in the orientation of the end-effector frame. We carried out kinesthetic demonstrations (Fig. \ref{fig:dustpan_setup}) and tracked the pose of the end-effector frame $E$ in world frame $O$. Thus, we generate the dataset $\mathscr{D}=\{\{(\mathbf{x}_{ij},\mathbf{p}_{ij},t_{ij})\}_{j=1}^{M_i}\}_{i=1}^{N}$, where $\mathbf{x}_{ij}\in\mathcal{S}^3/\mathbb{Z}_{2}$ is the quaternion representation of the orientation, $\mathbf{p}_{ij}\in\mathbb{R}^3$ is the position, and $t_{j}\in[0,1]$ is the normalized time. For this experiment, we considered $N = 12$ demonstrations, and we randomly sampled $M_i = 80$, $\forall i$, points from each recorded trajectory.

\begin{figure}[ht]
    \centering
    \includegraphics[width=0.47\textwidth]{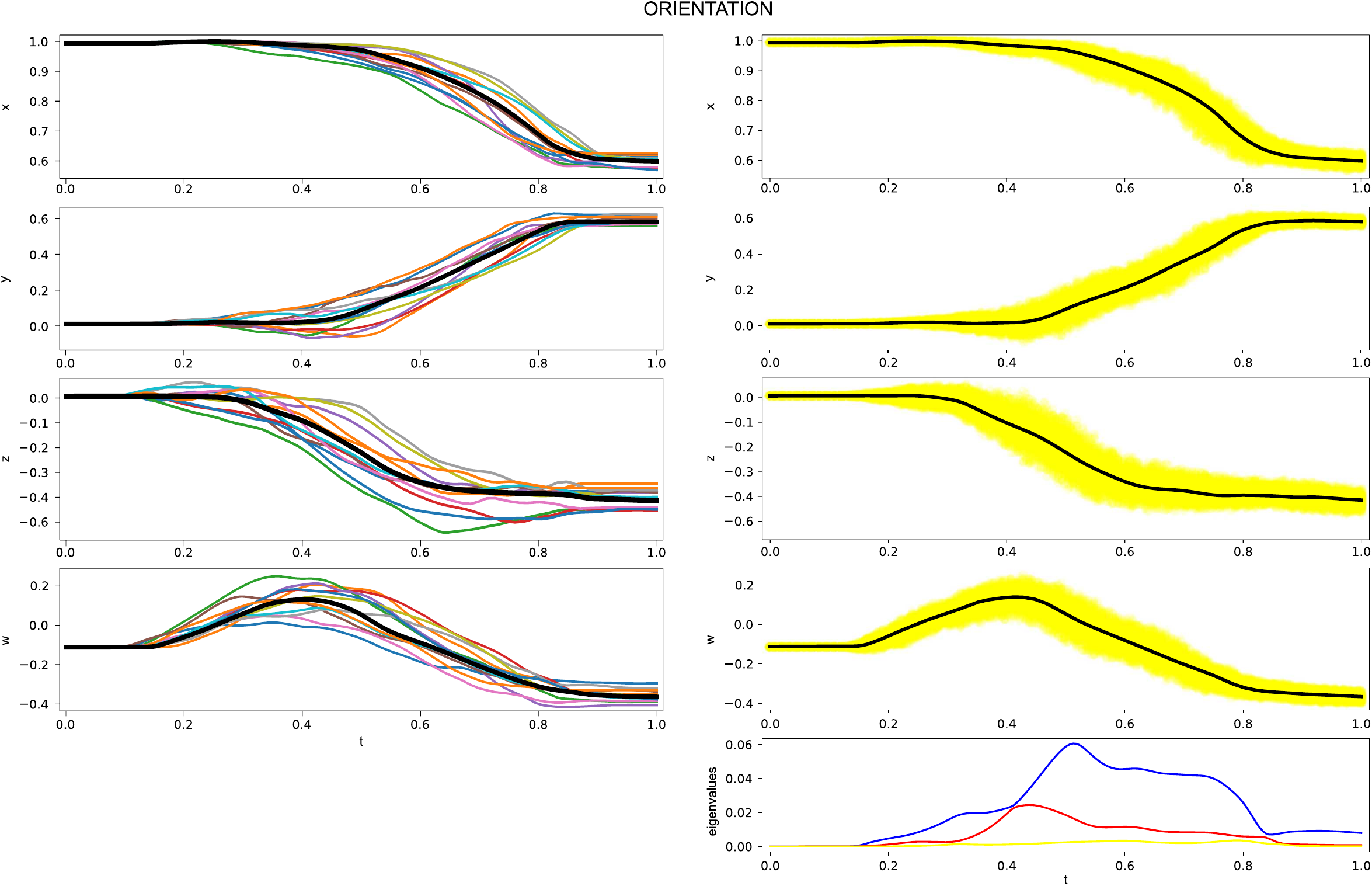}
    \caption{LEFT: Orientations of demonstrations $\mathscr{D}$, i.e. dataset $\{\{(\mathbf{x}_{ij},t_{ij})\}_{j=1}^{M_i}\}_{i=1}^{N}$, $\mathbf{x}_{ij}=(x_{ij},y_{ij},z_{ij},w_{ij})\in\mathcal{S}^{3}/\mathbb{Z}_2$. RIGHT: Mean $\hat{\mathbf{b}}_{(1)}(t)$ and density level by sampling, see text for details. RIGHT-BOTTOM: second through fourth eigenvalues of $\bm{\Lambda}(t)$}
    \label{fig:original_orientation}
\end{figure}

The left side of Fig. \ref{fig:original_orientation} shows the orientation set $\{\{(\mathbf{x}_{ij},t_{ij})\}_{j=1}^{M_i}\}_{i=1}^{N}$. Since $\mathbf{x}_{ij}$ is axial data, we ran a KLE algorithm with an ACG distribution as model. For 200 equidistant values of $t\in[0,1]$, we estimated the parameter $\hat{\bm{\Lambda}}(t)$ using the solution for KLE in Eq. (\ref{eq:acg_mle}). We used $h = 0.04$, and the Rao-Fisher metric (see, for example, Eq. (10) in \cite{rao_fisher}) to stop the iterations in Eq. (\ref{eq:acg_mle}) with a threshold of $1\times10^{-4}$. Due to the redundancy of $\bm{\Lambda}$, we normalised it so that $\mathrm{tr}(\hat{\bm{\Lambda}}(t))=d=4$, $\forall t$. The right-hand side of Fig. \ref{fig:original_orientation} shows the mean, $\hat{\mathbf{b}}_{(1)}(t)$, i.e. the eigenvector associated with the largest eigenvalue. In order to have a sense of density levels, for each $t$, $ACG(\hat{\bm{\Lambda}}(t))$ is simulated with 350 samples, then the 140 with the highest density (Eq. (\ref{eq:acg_def})) are plotted in Fig. \ref{fig:original_orientation}. It can be seen that the density follows the trend of the training data. 

\begin{figure}[ht]
    \centering
    \includegraphics[width=0.47\textwidth]{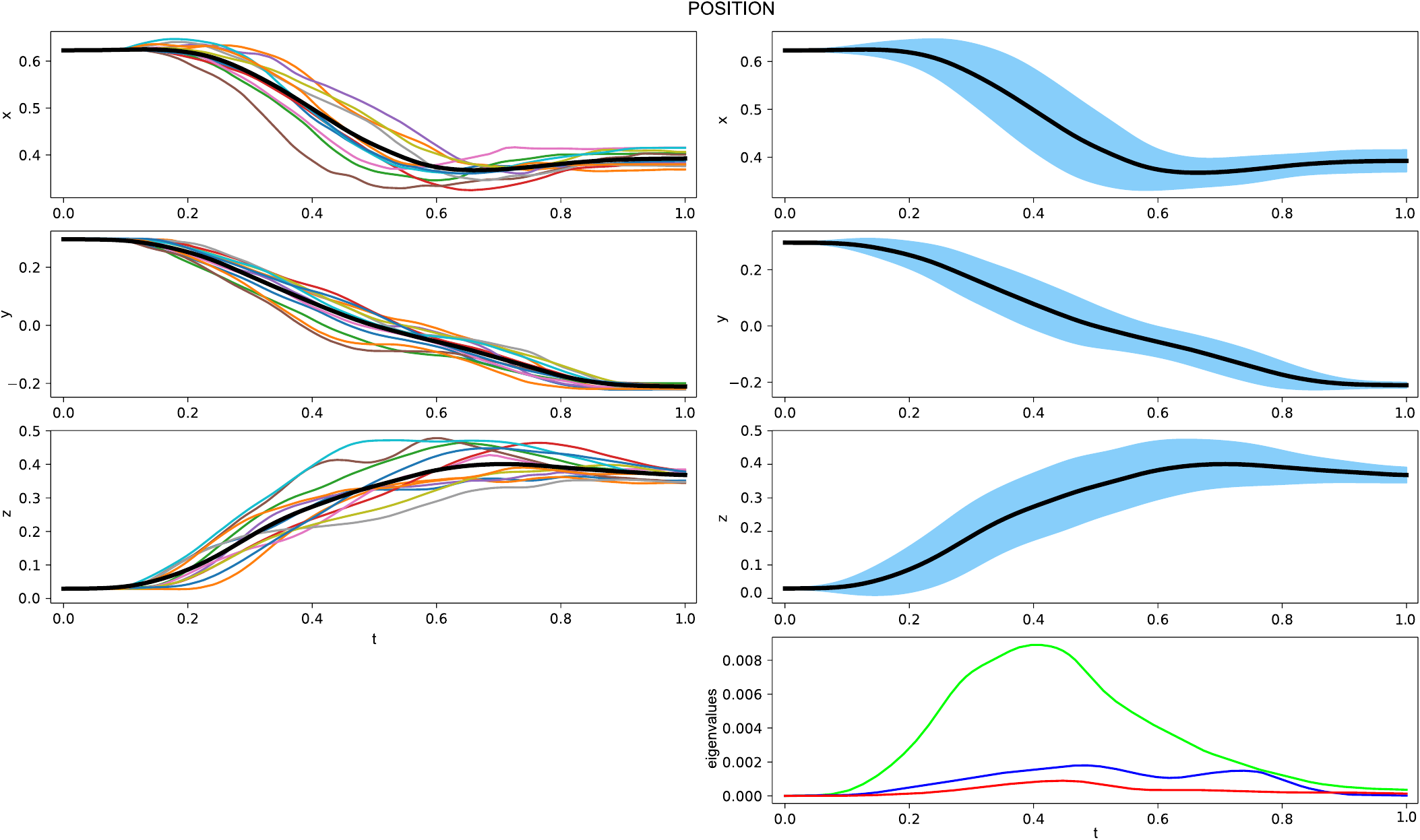}
    \caption{LEFT: Positions of demonstrations $\mathscr{D}$, i.e. dataset $\{\{(\mathbf{p}_{ij},t_{ij})\}_{j=1}^{M_i}\}_{i=1}^{N}$, $\mathbf{p}_{ij}=(x_{ij},y_{ij},z_{ij})\in\mathbb{R}^{3}$. RIGHT: Mean $\hat{\bm{\upmu}}(t)$ with 1.5 standard deviation density level. RIGHT-BOTTOM: eigenvalues of $\hat{\bm{\Sigma}}(t)$}
    \label{fig:original_position}
\end{figure}

The left side of Fig. \ref{fig:original_position} shows the position set $\{\{(\mathbf{p}_{ij},t_{ij})\}_{j=1}^{M_i}\}_{i=1}^{N}$. Since $\mathbf{p}_{ij}\in\mathbb{R}^{3}$, the task is much simpler and we ran a KLE algorithm with a multivariate Gaussian distribution as model. We used Eq. (\ref{eq:nw_normal}) to determine $\bm{\upmu}(t)$ and $\bm{\Sigma}(t)$, respectively. The results are shown on the right-hand side of Fig. \ref{fig:original_position}. 

\begin{figure}[ht]
    \centering
    \includegraphics[width=0.47\textwidth]{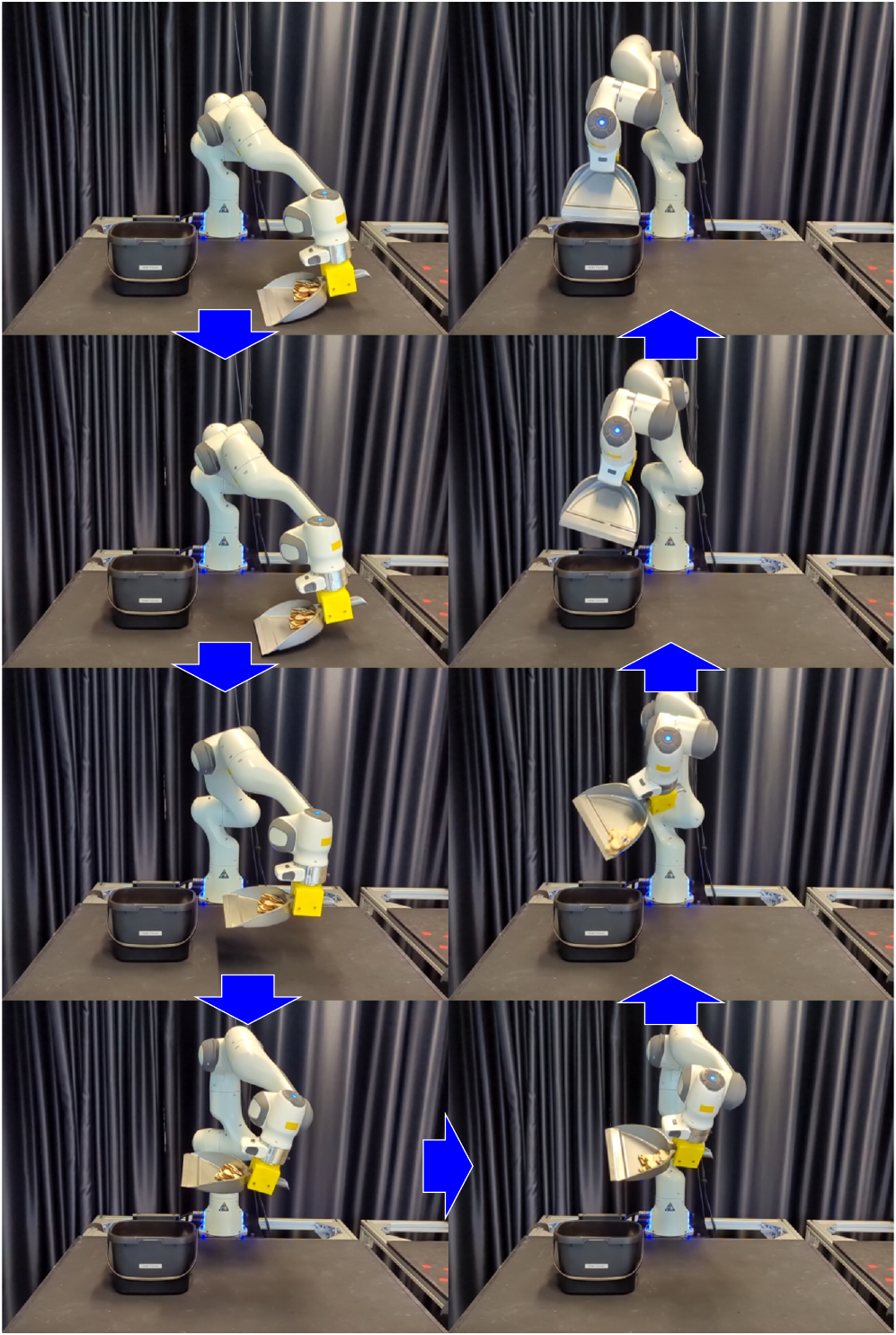}
    \caption{Franka Panda arm following the mean task-space trajectory obtained from $\mathscr{D}$. See video in the supplementary material.}
    \label{fig:video_mean}
\end{figure}

\begin{figure}[ht]
    \centering
    \includegraphics[width=0.47\textwidth]{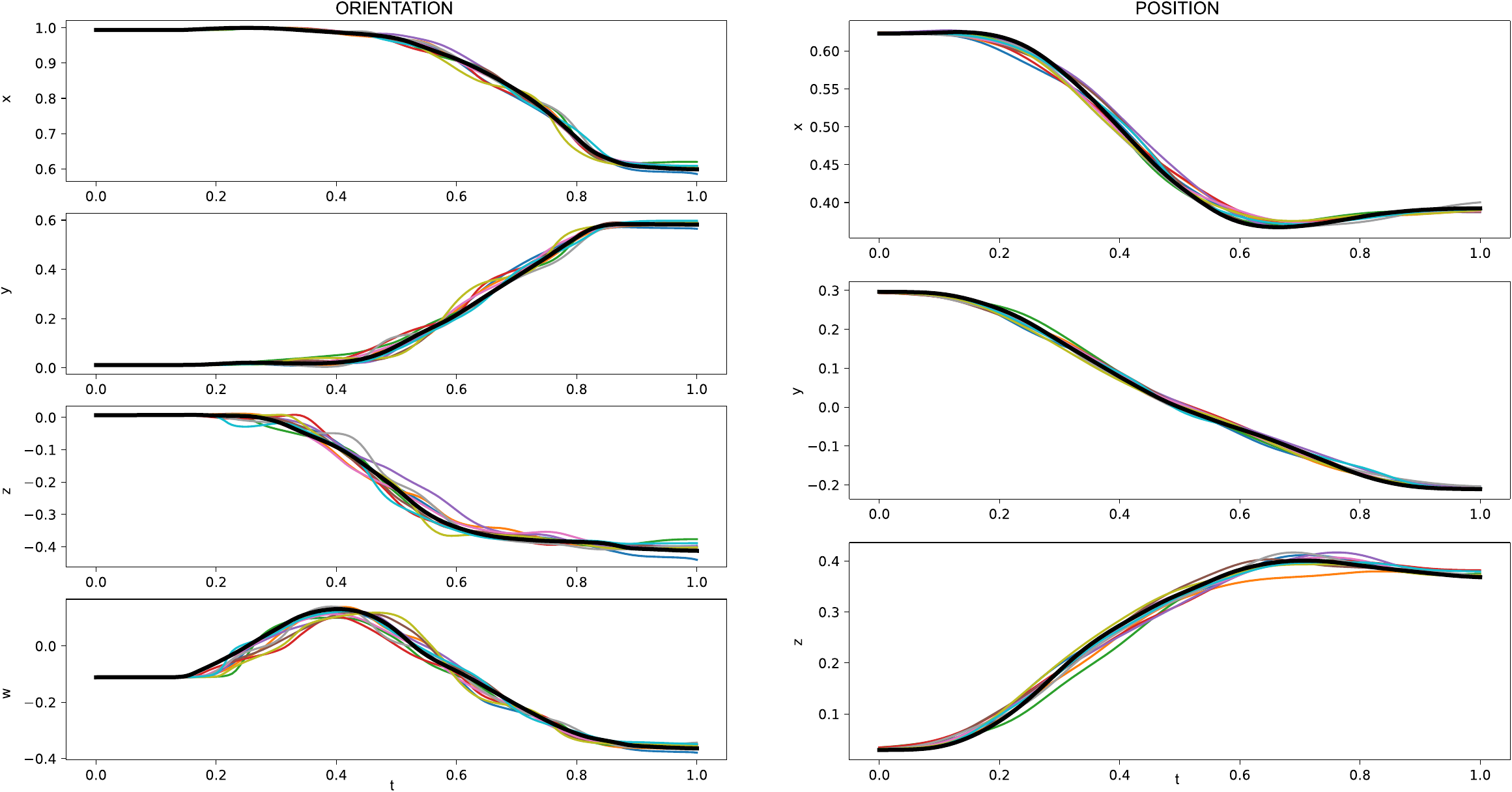}
    \caption{Ten trajectories generated from the learnt model using the method presented in Sec. \ref{sec:generation}}
    \label{fig:generated_curves}
\end{figure}

Fig. \ref{fig:video_mean} shows stills of the robot executing the mean trajectory $\{(\hat{\mathbf{b}}_{(1)}(t_k), \hat{\bm{\upmu}}(t_k), t_k)\}^{200}_{k=1}$. Fig. \ref{fig:generated_curves} shows several curves generated following the method proposed in Sec. \ref{sec:generation}. The curves are generated by smoothing through 120 points from the set $\{(\mathbf{x}_k, t_k)\}^{120}_{k=1}$, $\mathbf{x}_k\sim ACG(\hat{\bm{\Lambda}}(t_k))$, for orientation, and $\{(\mathbf{p}_k, t_k)\}^{120}_{k=1}$, $\mathbf{p}_k\sim \mathcal{N}(\hat{\bm{\upmu}}(t_k),\hat{\bm{\Sigma}}(t_k))$.

\corr{
It is now desired to adapt the learnt trajectories after the bin has been moved to a new pose. The final pose of the end-effector, ${}^{O}_{E}\mathbf{T}^{*}$, that will allow dropping the rubbish in the bin in its new pose at $t^{*}=1$ is known. It is known that frame $E$ must be at a new orientation with quaternion representation \corr{$\mathbf{x}^{*}=(0.7689, 0.4427, -0.2866, -0.3615)^{\top}$, and new position $\mathbf{p}^{*}=(0.45, -0.1, 0.4)$}. The desired concentration for the orientation part is governed by the eigenvalues matrix $\mathbf{A}^{*}=\mathrm{diag}(1\times10^4, 1, 1, 1)$. The desired parameter of the ACG distribution is then $\bm{\Lambda}^{*}=\mathbf{B}^{*}\mathbf{A}^{*}(\mathbf{B}^{*})^{\top}$, where $\mathbf{B}^{*}$ is obtained computing an orthonormal basis for the complement space of $\mathbf{x}^{*}$ to $\mathbb{R}^4$. On the other hand, the desired concentration for position is governed by $\bm{\Sigma}^{*}=1\times10^{-5}\mathbf{I}_{3}$.
}
\corr{
We followed the method presented in Sec. \ref{sec:adaptation}. For the orientation part we used activation functions $\alpha(t)=0.5-0.5\tanh(4.5(t-0.65))$. The dataset $\mathscr{E}=\{\{(\mathbf{y}_{i},t^{*})\}_{i=1}^{S}$ is generated by drawing $S=1.5N=120$ samples from $ACG(\bm{\Lambda}^{*})$. The bandwidth of the kernel for the points in $\mathscr{E}$ is set to $g=1$.
}
\corr{
The position part is done similarly with $S=N=80$ samples drawn from $\mathcal{N}(\mathbf{p}^{*}, \bm{\Sigma}^{*})$, and activation function $\alpha(t) = 0.5-0.5\tanh(10(t-0.6))$, and bandwith $g=0.3$ }

\begin{figure}[ht]
    \centering
    \includegraphics[width=0.49\textwidth]{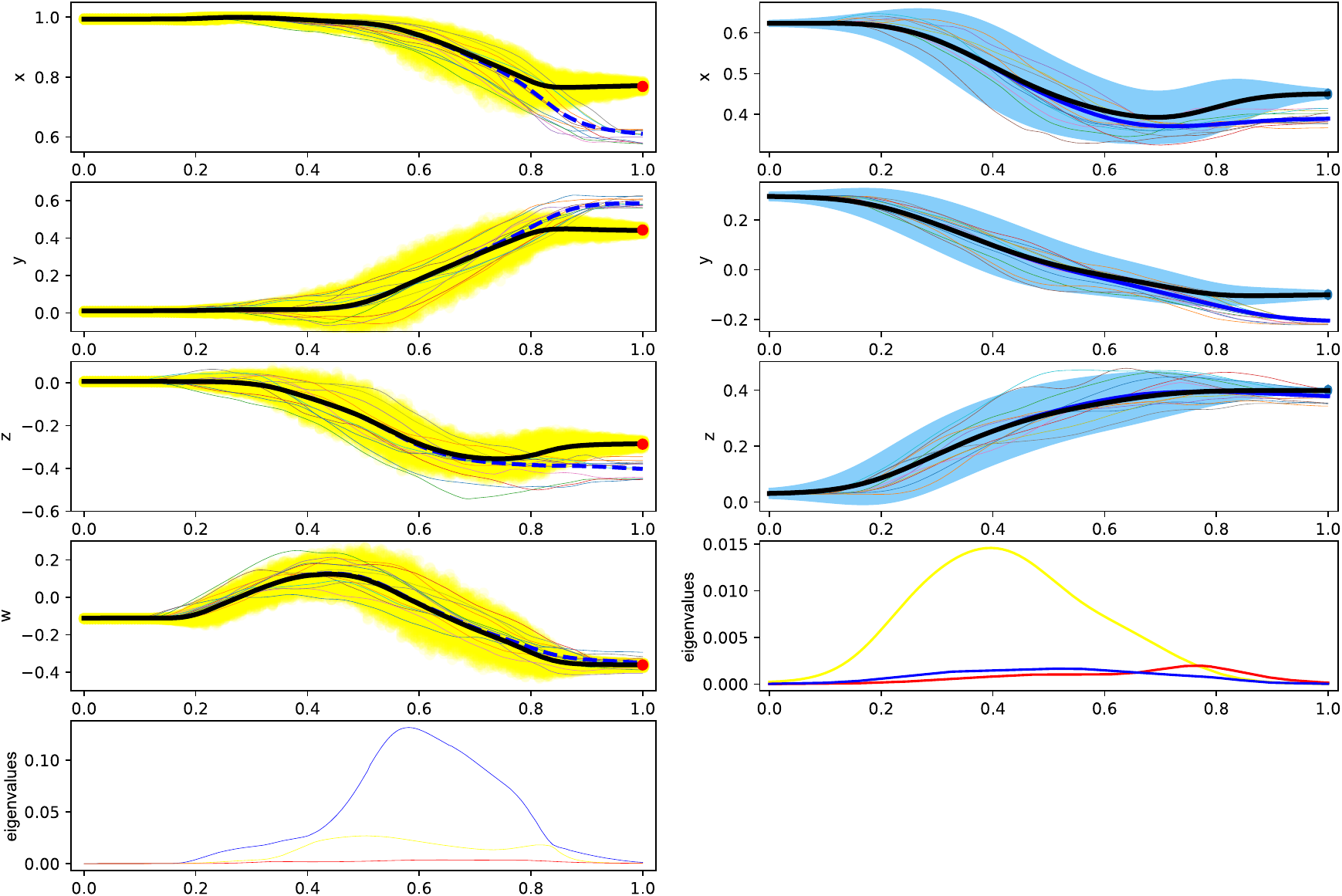}
    \caption{\corr{Adaptation of the learnt trajectories. $\mathbf{x}^{*}$ and $\mathbf{p}^{*}$ are shown in red dots. The adapted mean is shown in thick black curve. The yellow density levels of the orientation part are obtained by sampling, while the ones for position represent 1.5 covariance levels. The original trajectories of $\mathscr{D}$ are also shown as is their mean, in dashed thick curves. The bottom-left plot shows the second through fourth eigenvalues of $\bm{\Lambda}(t)$, while the bottom-right shows the eigenvalues of $\bm{\Sigma}(t)$}}
    \label{fig:adaptation_se3}
\end{figure}

\begin{figure}[ht]
    \centering
    \includegraphics[width=0.47\textwidth]{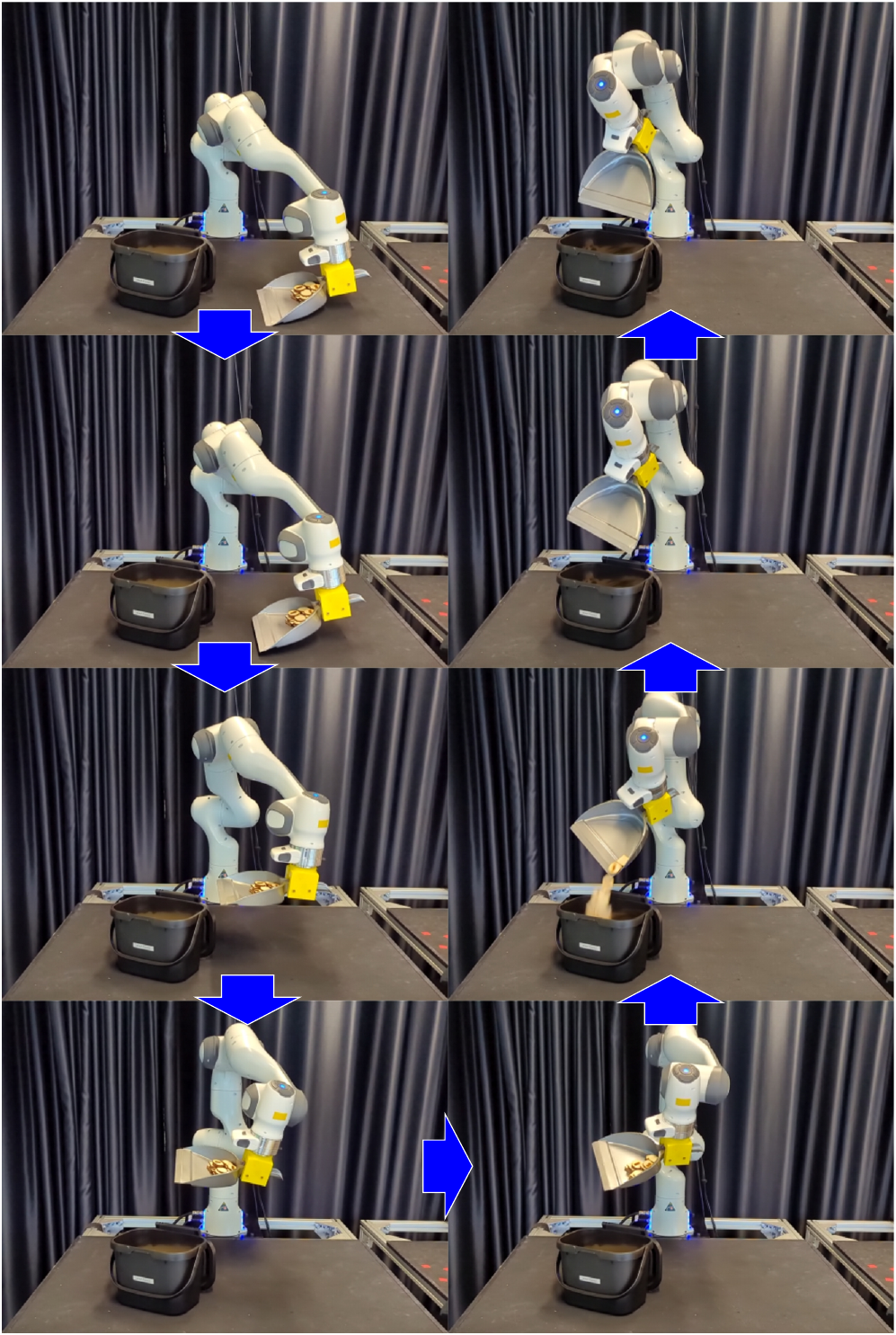}
    \caption{Franka Panda arm following the adapted task-space trajectory after the bin has been moved. See video in the supplementary material.}
    \label{fig:video_adapted}
\end{figure}

The results of this adaptation process are shown in Fig. \ref{fig:adaptation_se3}. It can be seen that the means achieve the desired new pose at the end of the trajectory. It can also be seen that the distributions for both position and orientation become highly concentrated at the end of the trajectory. Fig. \ref{fig:video_adapted} shows the robot following the adapted mean.

\begin{figure}[ht]
    \centering
    \includegraphics[width=0.49\textwidth]{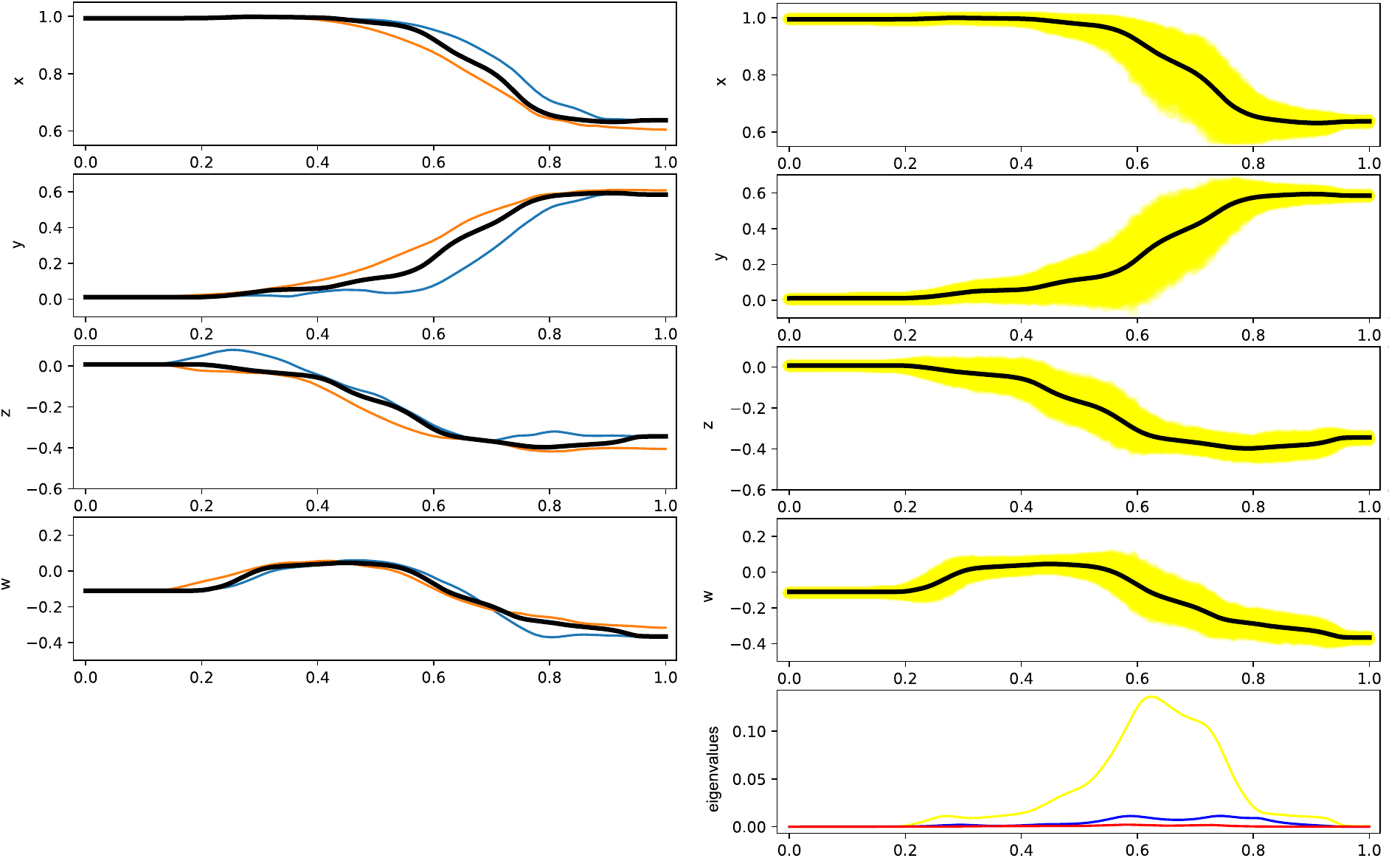}
    \caption{\corr{Model trained with only two trajectories (orientation component). LEFT: Demonstrated trajectories and estimated mean (thick black curve). RIGHT: Estimated mean and points sampled from the estimated distribution.}}
    \label{fig:minimal}
\end{figure}

\corr{Finally, in order to show the behaviour of the KLE method with a minimal dataset, Fig. \ref{fig:minimal} shows the results of training the orientation part of the model using only two trajectories. Since the distance between data points of different trajectories is larger, the bandwidth is changed to $h=0.9$.}

\subsection{Regression with response in $\mathrm{Sym}^{+}(6)$ and predictor in $\mathbb{R}^{2}\times\mathcal{S}^1$}\label{sec:exp_4}

In this experiment, simple regression (rather than trajectory learning) is carried out using the KLE method. The predictor is on a manifold, $\mathbb{R}^{2}\times\mathcal{S}^1$, which represents the position and orientation of a paint roller on a wall, while the response is in $\mathrm{Sym}^{+}(6)$, representing the manipulability ellipsoid of a demonstrator holding the paint roller. Such manipulability will be then transferred to the robot. This should be reflected in a more human posture. 

A 7-DOF Franka Emika Panda arm is to be taught to hold a paint roller against the wall with the manipulability that a human would impose. The demonstrations are a set of postures of a human holding a roller at different positions and orientations of the roller on a wall. For a new position and orientation of the roller, it is desired to predict the manipulability ellipsoid that the human would impose and, subsequently, transfer this to the robot. We assume that the angle that the handle of the roller makes with the wall is constant. Therefore, the predictor of this regression problem is in $\mathbb{R}^{2}\times\mathcal{S}^1$ (position and orientation of the roller on the wall) and the response is in $\mathrm{Sym}^{+}(6)$ (manipulability ellipsoid). Since, in this case, the rotation axis of the roller is perpendicular to its translation, $\mathbb{R}^{2}\times\mathcal{S}^1$ parametrises \emph{the group of planar displacements}, $\mathrm{P}(3)$ \cite{herveclass}. This is a 3-dimensional subgroup of $\mathrm{SE}(3)$ and is sometimes also referred to as $\mathrm{SE}(2)$.

Note that, unlike \cite{jaquier_spd} and \cite{luis_manipulability}, we are not interested in tracking manipulability from human to robot along a trajectory. Instead, we only aim to transfer the estimated manipulability for a single pose of the end-effector which puts the roller on the wall with a desired position and orientation. 

The training dataset for this experiment is $\mathscr{D}=\{(\mathbf{X}_{i},\mathbf{p}_i,\phi_i)\}_{i=1}^{N}$, where $\mathbf{X}_{i}\in\mathrm{Sym}^{+}(6)$, $\mathbf{p}_i\in\mathbb{R}^2$, $\phi_i\in\mathcal{S}^1$.

\begin{figure}[ht]
    \centering
    \includegraphics[width=0.47\textwidth]{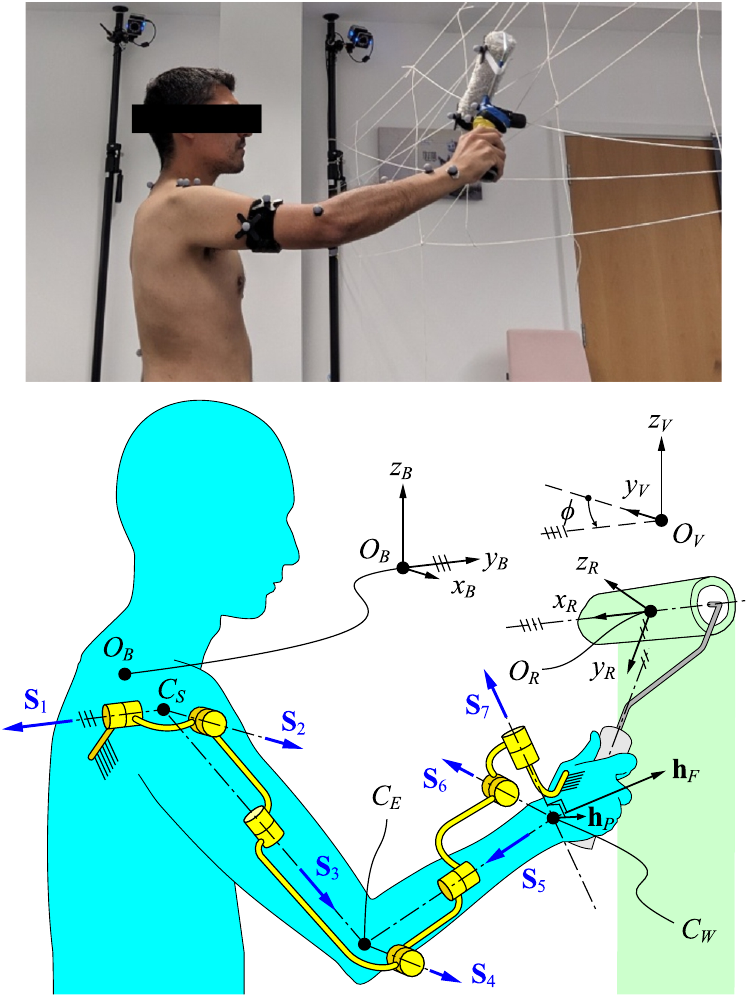}
    \caption{Data collection for experiment \ref{sec:exp_4}. TOP: A participant demonstrates how he holds a roller against a wall. The wall is a net of threads to allow the cameras track the markers. The markers set is defined for tracking of the joint centres, and thorax and roller frames. BOTTOM: Kinematic model and related coordinate systems for right arm holding a paint roller.}
    \label{fig:ex_spd6}
\end{figure}

We tracked the right arm of a participant using a Vicon System as shown in Fig. \ref{fig:ex_spd6} (top). For each demonstration $i$ we obtained the manipulability ellipsoid of the right arm from its Jacobian matrix as $\mathbf{X}_i=\mathbf{J}_{H,i}\mathbf{J}_{H,i}^{\top}$. To obtain $\mathbf{J}_{H,i}$, we use the kinematic model shown in Fig. \ref{fig:ex_spd6} (bottom). The model is a 7-DOF kinematic chain and was previously used in several publications \cite{arm_model_1,arm_model_2, muscle}. The Vicon world frame $V$ is fixed so that the wall coincides with the $y_V\text{--}z_V$ plane. Frame $B$ is the recommended by the International Society of Biomechanics \cite{isb} as thorax coordinate system. 

Let $\mathbf{S}_j$ be the screw axis of joint $j$, then the geometry of the kinematic chain modelling the arm is fully defined by the following constraints: $\mathbf{S}_1$, $\mathbf{S}_2$ and $\mathbf{S}_3$ intersect at $C_S$, the shoulder centre. Similarly, $\mathbf{S}_5$, $\mathbf{S}_6$ and $\mathbf{S}_7$ intersect at $C_W$, the wrist centre. $\mathbf{S}_4$ passes through $C_E$, the elbow centre. $\mathbf{S}_1$ is antiparallel to $y_B$, $\mathbf{S}_3$ is coincident with line $\overline{C_SC_E}$, and $\mathbf{S}_2$ is perpendicular to $\mathbf{S}_1$ and $\mathbf{S}_3$. $\mathbf{S}_5$ is coincident with line $\overline{C_EC_W}$, $\mathbf{S}_4$ is perpendicular to $\mathbf{S}_3$ and $\mathbf{S}_5$. $\mathbf{S}_7$ is perpendicular to $\mathbf{h}_F$ and $\mathbf{h}_P$, which are the directions of the middle finger, and the normal to the palm, respectively. Finally, $\mathbf{S}_6$ is perpendicular to $\mathbf{S}_5$ and $\mathbf{S}_7$.

If $\mathbf{S}_j\in\mathrm{se}(3)$ are the Pl\"ucker coordinates of the joint axes \cite{hunts_book}, the Jacobian matrix is given by $\mathbf{J}_{H} = [\mathbf{S}_1,\ldots,\mathbf{S}_7]\in\mathbb{R}^{6\times7}$. We track a frame $R$ fixed to the roller, as shown in Fig. \ref{fig:ex_spd6}. As claimed before, we assume that the angle between axis $y_R$ and the wall, i.e. plane $y_V\text{--}z_V$, is constant. We also assume that the distance between the shoulder centre, $C_S$, and the wall remains unchanged between demonstrations. Then, our predictor variables $(\mathbf{p},\phi)\in\mathbb{R}^2\times\mathcal{S}^1$ are defined as:
\begin{itemize}
    \item $\mathbf{p}\in\mathbb{R}^2$ is the projection of vector ${}^B\mathbf{r}_{O_R}-{}^B\mathbf{r}_{C_S}$ onto the wall (plane $y_V\text{--}z_V$).
    \item $\phi\in\mathcal{S}^1$ is the angle between $x_R$ and $y_V$ about $-x_V$.
\end{itemize}

A total of $N=80$ demonstrations were obtained. Then the KLE algorithm was used to predict a manipulability ellipsoid for a new pair $(\mathbf{p},\phi)\in\mathbb{R}^2\times\mathcal{S}^1$. This is done by modifying the metric in the kernel of Eq. (\ref{eq:nw_spd}). The estimator is then:

\[
    \hat{\mathbf{M}}(\mathbf{p},\phi)=\mathrm{exp}\left[\sum_{i=1}^{N}W_i(\mathbf{p},\phi)\mathrm{log}(\mathbf{X}_{i})\right],
\]

\noindent with
\[
    W_i(\mathbf{p},\phi):=\dfrac{K_h\Big(d_{\mathrm{P}(3)}\big((\mathbf{p}_i,\phi_i),(\mathbf{p},\phi)\big)\Big)}{\sum\limits_{i=1}^{N}K_h\Big(d_{\mathrm{P}(3)}\big((\mathbf{p}_i,\phi_i),(\mathbf{p},\phi)\big)\Big)},
\]

\noindent where $d_{\mathrm{P}(3)}(\cdot,\cdot)$ is a metric in $\mathrm{P}(3)$. Just like in $\mathrm{SE}(3)$, the lack of a natural length scale in $\mathrm{P}(3)$ does not allow for the introduction of a bi-invariant metric. Hence, as suggested in \cite{se3_metric_park}, a distance of weighted form is used:
\[
    d_{\mathrm{P}(3)}\left((\mathbf{p}_i,\phi_i),(\mathbf{p},\phi)\right):= w_{\phi}d_{\phi}(\phi_i,\phi)+w_{p}d_{p}(\mathbf{p}_i,\mathbf{p}),
\]

\noindent where $d_{\phi}(\cdot,\cdot)$ is a metric in $\mathcal{S}^1$ (smallest angle) and $d_{p}(\cdot,\cdot)$ is simply the Euclidean distance in $\mathbb{R}^2$. Similarly to \cite{se3_metric_averages}, we define such weights considering the averages of distances in the rotational and translational data to their respective means:

\begin{eqnarray*}
    w_{p} &:=& \left(\frac{1}{N}\sum\limits_{i=1}^{N}\|\overline{\mathbf{p}}_P-\mathbf{p}_i\|\right)^{-1},
    \\
    w_{\phi} &:=& \left(\frac{1}{N}\sum\limits_{i=1}^{N}d_{\phi}(\overline{\phi},\phi_i)\right)^{-1},
\end{eqnarray*}
\noindent where $\overline{\phi}$ is the mean for circular data in Eq. (2.2.4) of \cite{mardia_jupp}. Alternatively, \cite{se3_metric_chema} suggests $w_p=1$ and $w_{\phi}$ equal to the largest distance between two points of the projection of the rigid body onto the displacement plane. In our case, such a distance is the axial length of the roller, 0.23m. We tested this metric with very similar results. 

It is now necessary to transfer $\hat{\mathbf{M}}(\mathbf{p},\phi)$ to the robot. Since we are only interested in the proportion and orientation of the manipulability ellipsoid, we normalize the response as $\mathbf{X}_{H}:=\hat{\mathbf{M}}(\mathbf{p},\phi)/\lambda_{H,\mathrm{max}}$, where $\lambda_{H,\mathrm{max}}$ is the largest eigenvalue of $\hat{\mathbf{M}}(\mathbf{p},\phi)$.

Since ${}^{O}_{R}\mathbf{T}$, the pose of the roller frame $R$, is fully defined by a requested $(\mathbf{p},\phi)$, and since, once the roller is attached to the end-effector of the robot, the transformation ${}^{E}_{R}\mathbf{T}$ is known, then $(\mathbf{p},\phi)$ defines a unique pose of the end-effector, ${}^{O}_{E}\mathbf{T}={}^{O}_{R}\mathbf{T}(\mathbf{p},\phi)({}^{E}_{R}\mathbf{T})^{-1}$.

After a pose of the end-effector is defined by a requested $(\mathbf{p},\phi)$, the 1-DOF redundancy of the robot gives us one free parameter that allows us to adjust the manipulability ellipsoid of the Franka arm, which we normalise as $\mathbf{X}_{F}:=\mathbf{J}_F\mathbf{J}_F^{\top}/\lambda_{F,\mathrm{max}}$, where $\lambda_{F,\mathrm{max}}$ is the largest eigenvalue of $\mathbf{J}_F\mathbf{J}_F^{\top}$. $\mathbf{X}_{F}$ is optimised to be as close as possible to $\mathbf{X}_{H}$. To this end, we use the analytical solution for the inverse kinematics (IK) of the Franka Panda arm based on \cite{franka_ik}. This IK function has as input the end-effector pose, ${}^{O}_{E}\mathbf{T}$, and a value of the seventh joint angle, $q_7\in[q_{7,\mathrm{MIN}},q_{7,\mathrm{MAX}}]$, as free parameter. The function returns a maximum of 8 solutions $\mathbf{q}\in\mathbb{T}^7$. Therefore, for a given ${}^{O}_{E}\mathbf{T}$, we consider the Jacobian matrix as a function of $q_7$ alone. Thus, we are interested in the maximiser:
\begin{equation}\label{eq:q_7}
    q_7^{\#} = \arg\min_{q_7} d_{\mathrm{SPD}}\left(\mathbf{X}_{H},\,\mathbf{X}_{F}(q_7)\right),
\end{equation}
\noindent where $d_{\mathrm{SPD}}(\cdot,\cdot)$ is a metric in $\mathrm{Sym}^{+}(6)$. For these experiments, we chose the Rao-Fisher metric. 

Since this is a 1-dimensional bounded search, the optimization in Eq. (\ref{eq:q_7}) can be simply done by discretising $[q_{7,\mathrm{MIN}}$, $\,q_{7,\mathrm{MAX}}]$, calculating the IK for each angle and picking the one that minimizes (\ref{eq:q_7}). Since our IK is analytical, this process much faster than what it would be if we used any of the numerical IK solvers, which are the commonplace for 7-DOF robots.

We ran the experiment for a requested pose of the roller $(\mathbf{p},\phi)=((-0.20,0.55), \pi/6)$. We set $h=0.26$ from cross-validation. The eigenvalues and eigenvectors of the estimated $\mathbf{X}_H$ and transferred $\mathbf{X}_F$ are shown in \ref{sec:appendix_eigen}.

\begin{figure}[ht]
    \centering
    \includegraphics[width=0.47\textwidth]{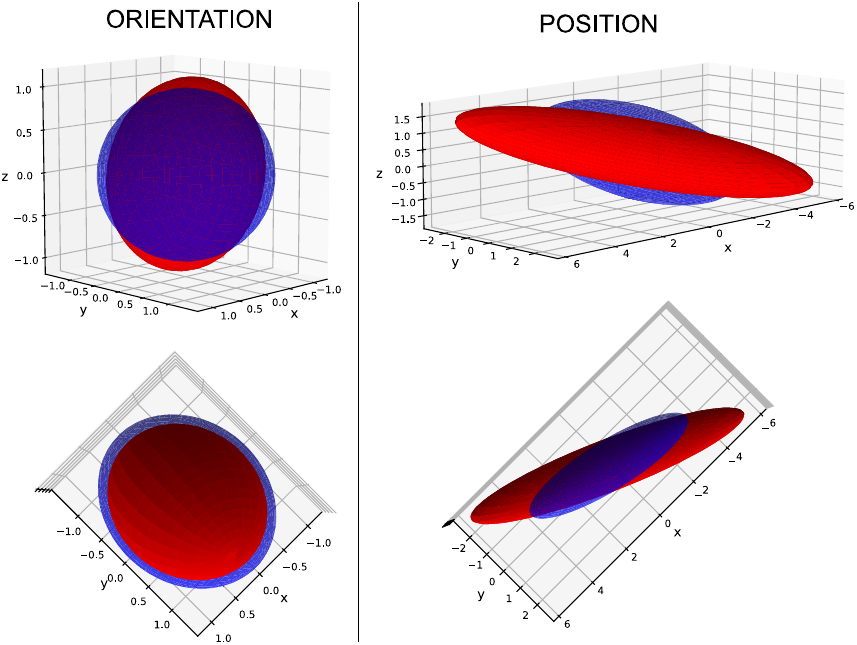}
    \caption{Position and orientation ellipsoids for $\mathbf{X}_H$ (red) and $\mathbf{X}_F$ (blue)}
    \label{fig:ellipsoids_plots}
\end{figure}

Eq. (\ref{eq:q_7}) returned $q_7^{\#}=1.412$, with $\mathbf{q}^{\#} = (-0.886, 1.651, 2.154, -1.528, 2.322, 2.165, 1.412)$ as whole configuration of the arm. The corresponding position and orientation ellipsoids for $\mathbf{X}_F$ and $\mathbf{X}_H$ are shown in Fig. \ref{fig:ellipsoids_plots}.

\begin{figure}[ht]
    \centering
    \includegraphics[width=0.47\textwidth]{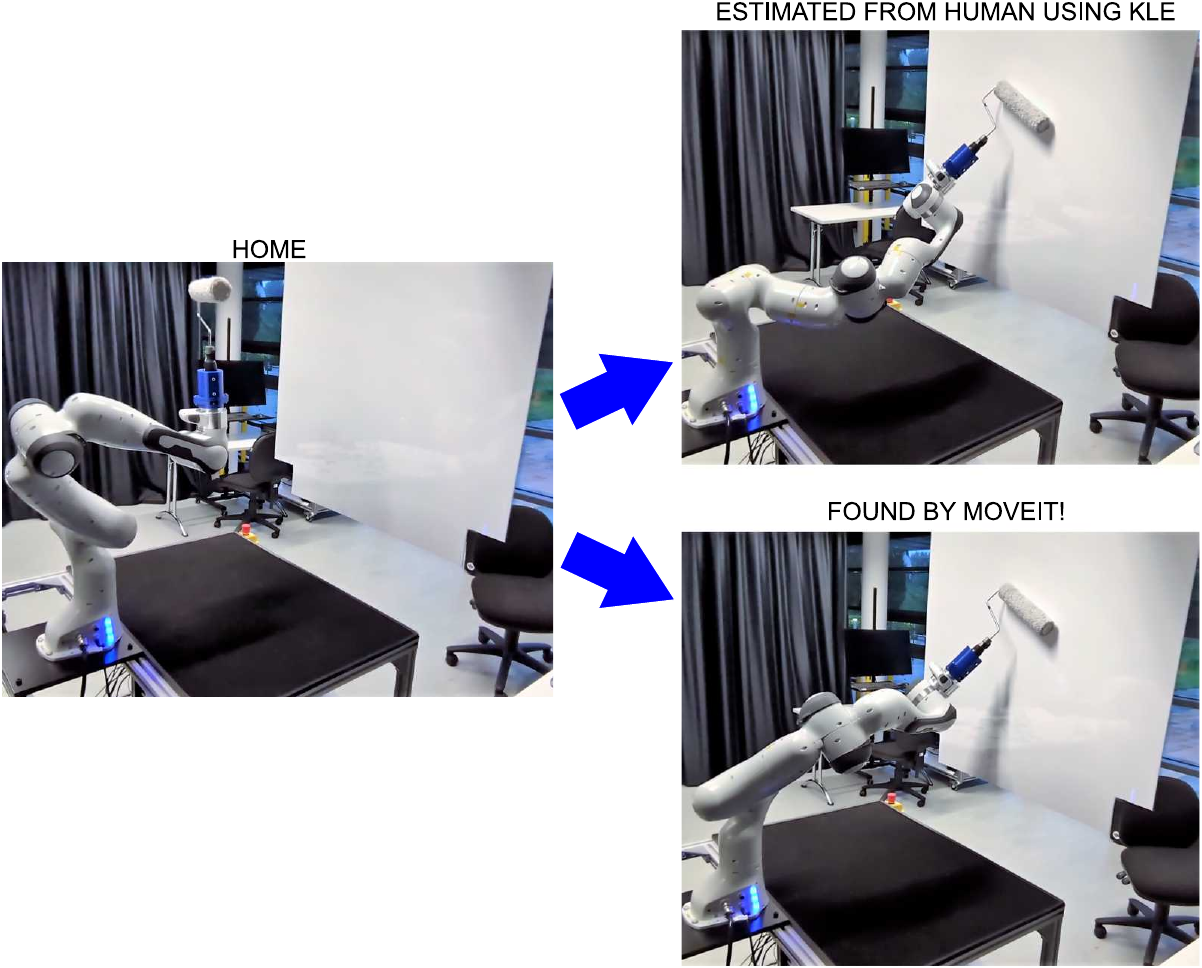}
    \caption{The Franka Panda arm is initially in a home configuration $\mathbf{q}_0$ (LEFT). Then $\mathbf{q}^{\#}$, which optimises $\mathbf{X}_F$ to be as close as possible to $\mathbf{X}_H(\mathbf{p},\phi)$ is set as as {\it joint goal} on Moveit! This takes the robot to $\mathbf{q}^{\#}$ (RIGHT-TOP). Starting again from $\mathbf{q}_0$, ${}^{O}_{E}\mathbf{T}(\mathbf{p},\phi)$ is set as {\it pose goal} on Moveit! which takes the robot to the configuration shown on the RIGHT-BOTTOM. See video in the supplementary material.}
    \label{fig:ellipsoids_photos}
\end{figure} 

Fig. \ref{fig:ellipsoids_photos} shows the Franka Panda arm in a home configuration, $\mathbf{q}_0$, from which it moves to the pose ${}^{O}_{E}\mathbf{T}$ defined by the requested $(\mathbf{p},\phi)$. We used MoveIt! to plan from $\mathbf{q}_0$ to the computed $\mathbf{q}^{\#}$ (Fig. \ref{fig:ellipsoids_photos}, right-top). As a comparison, we then asked MoveIt! to plan from the same home configuration to ${}^{O}_{E}\mathbf{T}(\mathbf{p},\phi)$ (Fig. \ref{fig:ellipsoids_photos}, right-bottom). When compared to the final posture achieved by the task-space MoveIt! planner, it can be seen that the posture $\mathbf{q}^{\#}$ obtained with our method better imitates the posture that a human would have when holding the paint roller against the wall. 


\section{Conclusions}\label{sec:conclusions}

In this paper, the method of kernelised likelihood estimation (KLE) was proposed as a tool for on-manifold LfD. The method is based on the idea of local likelihood, where the terms of the log-likelihood expression of a probability density function are weighted with kernels. The result is a non-parametric regression on the parameters of the probability distribution. Since this concept is independent of the probability model, the learning method can be applied on any manifold as long as the selected probability model has a solution for the maximum likelihood estimation (MLE) for i.i.d. data. This generality allowed us to apply the method to trajectory learning on different manifolds. The learning is probabilistic, meaning that, for a new value of the predictor, we are able to estimate not only the mean of the response, but also the dispersion structure of it. Unlike most of the trajectory learning methods in robotics, KLE works intrinsically on the manifold, rather than on projections on one or several tangent spaces. 

\corr{In the first experiment (Sec. \ref{sec:exp_2}),} the mean curve by the \corr{KLE} method crosses more centrally the data compared to the GMR \corr{and KMP methods.} This experiment also showed that he KLE method is applicable even when no simple solution for the MLE is known. In which case Eq. (\ref{eq:weighted_mle}) has to be solved directly with a numerical solver. In our second experiment (Sec. \ref{sec:exp_2}), we used the same probability model employed by GMR, the SPD log-normal distribution. However, our method recovered the mean of the ground truth more precisely than GMR. Our method was also able to recover the covariance matrices of the ground truth. The distortion seen in the results obtained with GMR are a consequence of the projection onto tangent spaces.

In our third experiment (Sec. \ref{sec:exp_3}), the KLE method was applied to the learning pose trajectories with a large change of orientation. The robot learnt to use trajectories for dropping rubbish into a bin using a dustpan. Adaptation of the trajectories to a new position of the dustpan was achieved. The fourth experiment (Sec. \ref{sec:exp_4}) shown the application of the KLE method as a regression tool with both predictor and responds lying on manifolds. The task was to teach a redundant robot arm to hold a paint roller against the wall with the manipulability that a human would impose to it. 

It is commonly thought that the use of intrinsic probability distributions for manifold-valued data results in intractable algorithms for robot learning. This belief probably stems from the complexity of the normalising term of the Bingham distribution, whose computation is normally done through look-up tables or even neural networks \cite{bingham_rl}. In this paper, we used the ACG distribution as an alternative to the Bingham distribution. \corr{The considerations for selecting an adequate model are the same as for the IID case: which is more convenient on the grounds of (i) ease of simulation and (ii) ease of evaluating the likelihood (especially in relation to tractability of normalising constant), and (iii) which is better fitting to the data. See \cite{esag} and \cite{tyler_acg} for discussion of these trade-offs. For the ESAG vs Kent \cite{kent} (for directional data), and ACG vs Bingham (for axial data) examples, there is little to choose between the distributions on the grounds of (iii), and we favour ESAG and ACG respectively on grounds of (i) and (ii). }

\corr{
The benefits of the KLE method come at the cost of the following limitations which can encourage future work:
\begin{itemize}
    \item The choice of parameters in an adaptation task largely affects the result. With the activation functions used in this paper, the number of parameters is four. In \ref{sec:appendix_adaptation}, an analysis of the effects of these parameters was presented. However, a deeper study on the selection of those is needed.
    \item The running times are longer. However, our code was developed in Python without focusing on the speed performance.
    \item The method method proposed in Sec. \ref{sec:generation} for the generation of trajectories based on the learnt model provides curves that tend to fluctuate around the mean. The generation of random trajectories has to be improved so that the dispersion of sampled curves better reflects the variability of the original demonstrations.
\end{itemize}
}

In addition to the above-mentioned opportunities of improvement, in future research, we will apply the KLE method to other types of manifold-valued data that were not covered in this paper, for example the Grassmannian manifold. 


\section*{Acknowledgment}
\addcontentsline{toc}{section}{Acknowledgment}
This work was supported by UK Research and Innovation (UKRI), CHIST-ERA (HEAP: EP/S033718/2), National Science Foundation (NSF DMS-2015374), and the National Institute of Health (NIH R37-CA214955).


\bibliographystyle{IEEEtran}
\bibliography{bibliography_nonparametric.bib} 

\appendices

\section{Proofs} \label{sec:appendix}

\noindent {\bf Proof of Eq. (\ref{eq:nw_normal}).} The log-likelihood function for a $\mathcal{N}(\bm{\upmu},\bm{\Sigma})$-distributed observation, $\mathbf{x}_i$, is

\begin{equation}\label{app:eq:loglik}
    \ell_i=-\frac{1}{2}\log|2\pi\bm{\Sigma}|-\frac{1}{2} (\mathbf{x}_i-\bm{\upmu})^{\top}\bm{\Sigma}^{-1}(\mathbf{x}_i-\bm{\upmu}),
\end{equation}
and for an independent sample 
$\{\mathbf{x}_i\}_{i=1,...,N}$ of such observations, the log-likelihood, 
$\ell=\sum_{i=1}^{N}\ell_i$, can be written \cite{mkb}
\begin{eqnarray}\label{app:eq:loglik_equiv}
    \ell &=& -\frac{N}{2}\log|2\pi\bm{\Sigma}| -\frac{N}{2}\mathrm{tr}\left(\bm{\Sigma}^{-1}\mathbf{S}\right)
    \nonumber\\&{}&
    -\frac{N}{2}(\overline{\mathbf{x}}-\bm{\upmu})^{\top}\bm{\Sigma}^{-1}(\overline{\mathbf{x}}-\bm{\upmu}),
\end{eqnarray}
where $\overline{\mathbf{x}}=N^{-1}\sum_{i=1}^N\mathbf{x}_i$ and $\mathbf{S}=N^{-1}\sum_{i=1}^N(\mathbf{x}_i-\bm{\upmu})(\mathbf{x}_i-\bm{\upmu})^{\top}$. From (\ref{app:eq:loglik_equiv}) it follows (see e.g. \cite{mkb}) that the MLEs of the model parameters are $\hat{\bm{\upmu}}=\overline{\mathbf{x}}$ and $\hat{\bm{\Sigma}}=\mathbf{S}$.

We consider now the $\bm{\upmu}$ and 
${\bm{\Sigma}}$ that maximise the 
weighted log-likelihood, $\tilde{\ell} =\sum_{i=1}^{N}W_i\ell_i$, where $W_i\geq 0$ and $\sum_{i=1}^{N}W_i = 1$. We have
\begin{equation}\label{app:eq:weighted_loglik}
    \tilde{\ell} = -\frac{1}{2}\log|2\pi\bm{\Sigma}|-\frac{1}{2}\sum_{i=1}^NW_i(\mathbf{x}_i-\bm{\upmu})^{\top}\bm{\Sigma}^{-1}(\mathbf{x}_i-\bm{\upmu}),
\end{equation}
and will define
\[
    \tilde{\mathbf{x}}=\sum_{i=1}^NW_i\mathbf{x}_i, \qquad
    \tilde{\mathbf{S}}=\sum_{i=1}^NW_i(\mathbf{x}_i-\tilde{\mathbf{x}})(\mathbf{x}_i-\tilde{\mathbf{x}})^{\top},
\]
as weighted analogues of $\overline{\mathbf{x}}$ and $\mathbf{S}$, respectively. Using the identity
\begin{align*}
    & (\mathbf{x}_i-\bm{\upmu})^{\top}\bm{\Sigma}^{-1}(\mathbf{x}_i-\bm{\upmu})
    \\
 = & \, (\mathbf{x}_i-\tilde{\mathbf{x}})^{\top}\bm{\Sigma}^{-1}(\mathbf{x}_i-\tilde{\mathbf{x}})
    +(\tilde{\mathbf{x}}-\bm{\upmu})^{\top}\bm{\Sigma}^{-1}(\tilde{\mathbf{x}}-\bm{\upmu}) \\
    & + 2(\tilde{\mathbf{x}}-\bm{\upmu})^{\top}\bm{\Sigma}^{-1}(\mathbf{x}_i-\tilde{\mathbf{x}}),
\end{align*}
thus
\begin{align*}
    & \sum_{i=1}^{N}W_i(\mathbf{x}_i-\bm{\upmu})^{\top}\bm{\Sigma}^{-1}(\mathbf{x}_i-\bm{\upmu}) \\ 
    = & \sum_{i=1}^{N}W_i(\mathbf{x}_i-\tilde{\mathbf{x}})^{\top}\bm{\Sigma}^{-1}(\mathbf{x}_i-\tilde{\mathbf{x}})
    + (\tilde{\mathbf{x}}-\bm{\upmu})^{\top}\bm{\Sigma}^{-1}(\tilde{\mathbf{x}}-\bm{\upmu}),
\end{align*}
and by writing $(\mathbf{x}_i-\tilde{\mathbf{x}})^{\top}\bm{\Sigma}^{-1}(\mathbf{x}_i-\tilde{\mathbf{x}})$ $=\mathrm{tr}\big(\bm{\Sigma}^{-1}(\mathbf{x}_i-\tilde{\mathbf{x}})(\mathbf{x}_i-\tilde{\mathbf{x}})^{\top}\big)$ it follows that\[
    \mathrm{tr}\left\{\bm{\Sigma}^{-1}\left[\sum_{i=1}^NW_i(\mathbf{x}_i-\tilde{\mathbf{x}})(\mathbf{x}_i-\tilde{\mathbf{x}})^{\top}\right]\right\} = \mathrm{tr}\big(\bm{\Sigma}^{-1}\tilde{\mathbf{S}}\big).
\]
Hence Eq. (\ref{app:eq:weighted_loglik}) becomes
\begin{eqnarray*}
    \tilde{\ell} &=& -\frac{1}{2}\log|2\pi\bm{\Sigma}|-\frac{1}{2}\mathrm{tr}\big(\bm{\Sigma}^{-1}\tilde{\mathbf{S}}\big)
    \\&{}&
    -\frac{1}{2}(\tilde{\mathbf{x}}-\bm{\upmu})^{\top}\bm{\Sigma}^{-1}(\tilde{\mathbf{x}}-\bm{\upmu}),
\end{eqnarray*}
which is a direct analogue of Eq. (\ref{app:eq:loglik_equiv}), except with $\mathbf{S}$ replaced by $\tilde{\mathbf{S}}$, $\overline{\mathbf{x}}$ replaced by $\tilde{\mathbf{x}}$, and differing by the constant factor $N$; and consequently it is maximised by
\[
    \hat{\bm{\upmu}} = \tilde{{\mathbf{x}}},\qquad \hat{\bm{\Sigma}} = \tilde{\mathbf{S}}.
\]
$\hfill\blacksquare$


\noindent {\bf Proof of Eq. (\ref{eq:acg_kernel_lik}).} The log-likelihood for an $ACG(\bm{\Lambda})$-distributed observation, $\mathbf{x}_i$, is:
\begin{equation}\label{app:eq:acg_loglik}
    \ell_i = \log\Gamma\left(\frac{d}{2}\right)-\log 2\sqrt{\pi^d|\bm{\Lambda}|}-\frac{d}{2}\log\mathbf{x}_i^{\top}\bm{\Lambda}^{-1}\mathbf{x}_i   
\end{equation}

Therefore, with rearrangement of the second term of (\ref{app:eq:acg_loglik}), the weighted log-likelihood of for a set $\{\mathbf{x}_i\}_{i=1}^N$ is:

\begin{eqnarray}\label{app:eq:acg_loglik_weighted}
    \tilde{\ell} &=& \log\Gamma\left(\frac{d}{2}\right)-\log 2\pi^{\frac{d}{2}}-\frac{1}{2}\log|\bm{\Lambda}|
    \nonumber\\&{}&
    -\frac{d}{2}\sum_{i=1}^NW_i\mathrm{log}\,\mathbf{x}_i^{\top}\bm{\Lambda}^{-1}\mathbf{x}_i
\end{eqnarray}

We wish to obtain the maximiser of $\tilde{\ell}$, thus we set $\partial\tilde{\ell}/\partial\bm{\Lambda}=\mathbf{0}$, the $d$-by-$d$ matrix of zeros. From \cite{cookbook}, partial derivation of the third term of (\ref{app:eq:acg_loglik_weighted}) gives:
\begin{equation}\label{app:eq:deriv_4}
    \frac{\partial}{\partial\bm{\Lambda}}\log|\bm{\Lambda}| = \left(\bm{\Lambda}^{\top}\right)^{-1}=\bm{\Lambda}^{-1},
\end{equation}
\noindent while for the contribution of the $i$th observation to the fourth term of (\ref{app:eq:acg_loglik_weighted}) we have:
\begin{eqnarray}\label{app:eq:deriv_5}
    \frac{\partial}{\partial\bm{\Lambda}}\mathbf{x}_i^{\top}\bm{\Lambda}^{-1}\mathbf{x}_i = \left(\mathbf{x}_i^{\top}\bm{\Lambda}^{-1}\mathbf{x}_i\right)^{-1}\frac{\partial}{\partial\bm{\Lambda}}\left(\mathbf{x}_i^{\top}\bm{\Lambda}^{-1}\mathbf{x}_i\right)
    &{}&\nonumber\\
    =-\left(\mathbf{x}_i^{\top}\bm{\Lambda}^{-1}\mathbf{x}_i\right)^{-1}\bm{\Lambda}^{-\top}\mathbf{x}_i\mathbf{x}_i^{\top}\bm{\Lambda}^{-\top}.&{}&
\end{eqnarray}
From (\ref{app:eq:deriv_4}) and (\ref{app:eq:deriv_5}), it follows that:
\[
    \frac{\partial\tilde{\ell}}{\partial\bm{\Lambda}}=-\frac{1}{2}\bm{\Lambda}^{-1}+\frac{d}{2}\sum_{i=1}^NW_i\left(\mathbf{x}_i^{\top}\bm{\Lambda}^{-1}\mathbf{x}_i\right)^{-1}\bm{\Lambda}^{-\top}\mathbf{x}_i\mathbf{x}_i^{\top}\bm{\Lambda}^{-\top}.
\]
Hence, by setting $\partial\tilde{\ell}/\partial\bm{\Lambda}=0$, it follows that:
\begin{eqnarray*}
    \bm{\Lambda}^{-1}&=&d\sum_{i=1}^NW_i\frac{\bm{\Lambda}^{-1}\mathbf{x}_i\mathbf{x}_i^{\top}\bm{\Lambda}^{-1}}{\mathbf{x}_i^{\top}\bm{\Lambda}^{-1}\mathbf{x}_i}
    \\
    \Rightarrow \bm{\Lambda} &=&d\sum_{i=1}^NW_i\frac{\mathbf{x}_i\mathbf{x}_i^{\top}}{\mathbf{x}_i^{\top}\bm{\Lambda}^{-1}\mathbf{x}_i}.
\end{eqnarray*}
$\hfill\blacksquare$

\noindent {\bf Proof of Eq. (\ref{eq:nw_spd}).} Since the SPD log-normal distribution is an analog of the multivariate normal distribution with vectorised versions of the logarithms of SPD data, Eq. (\ref{eq:nw_spd}) follows directly from the proof of Eq.(\ref{eq:nw_normal})
$\hfill\blacksquare$


\vspace{0.5cm}
\noindent {\bf Proof of claim in Sec. \ref{sec:generation} about relation to GPs.}
Consider the scalar case $\mathcal{M}=\mathbb{R}$ and $t\in\mathbb{R}$ with $P_{\mathcal{M}}(\theta)=\mathcal{N}(\mu,\sigma^2)$, $\theta=\{\mu,\sigma^2\}$, and let $\mathbf{x}_r=(x_{r,1},\ldots,x_{r,N_r})^{\top}$, $\hat{\bm{\upmu}}=\big(\hat{\mu}(t_{r,1}),\ldots,\hat{\mu}(t_{r,N_r})\big)^{\top}$. Then $\mathbf{x}_r\sim\mathcal{N}(\hat{\mu},\hat{\sigma}^2\mathbf{I})$. Let $\mathbf{t}=(t_1,\ldots,t_{N_r})$ be chosen times of which to evaluate $\hat{\mu}_r(t)$ and write $\hat{\bm{\upmu}}_r=\big(\hat{\mu}_r(t_1),\ldots,\hat{\mu}_r(t_{N_r})\big)$. Then, from (\ref{eq:scalar}), $\hat{\bm{\upmu}}_r=\mathbf{S}\mathbf{x}_r$, where $\mathbf{S}$ is the matrix whose $(i,j)$ element, $S_{ij}$, per (\ref{eqn:weights:defn}), equals:
\[
    S_{ij} = \frac{K_h(t_i-t_{r,j})}{\sum^{N_r}_{j=1}K_h(t_i-t_{r,j})},
\]
hence, $\hat{\bm{\upmu}}_r$ is normally distributed for any choice of $\mathbf{t}_{r}$. Thus, $\hat{{\upmu}}_r(t)$ is a GP. For the case with $\mathcal{M}=\mathbb{R}^d$, $d>1$ and $P_{\mathcal{M}}(\theta)$ being the multivariate normal distribution, similar arguments show that $\hat{\bm{\upmu}}_r({t})$ is a GP. 
$\hfill\blacksquare$

\section{Effects of the choice of parameters in an adaptation task} \label{sec:appendix_adaptation}

\begin{figure}[ht]
    \centering
    \includegraphics[width=0.47\textwidth]{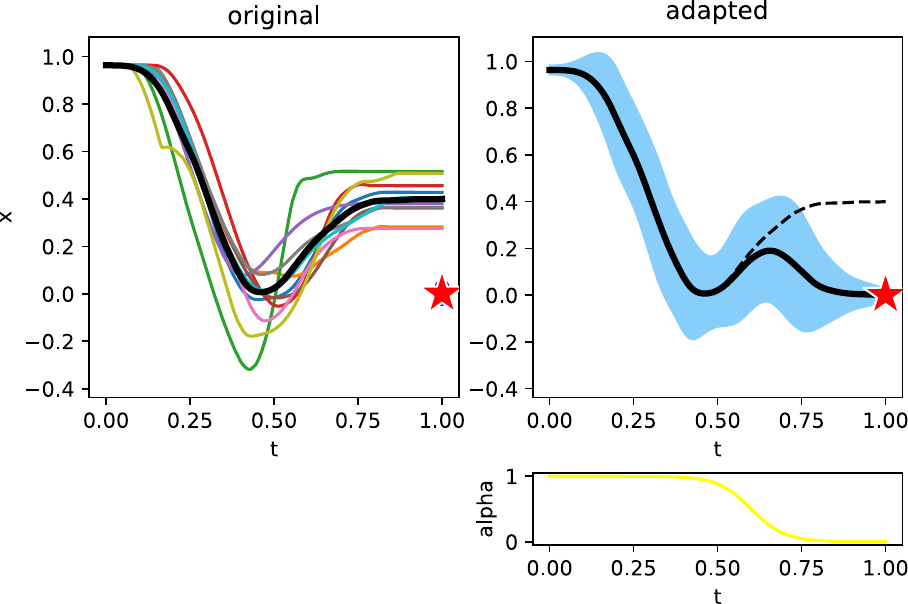}
    \caption{\corr{Adaptation of a scalar model to a desired final point that is outside the training data. The data are $N=10$ trajectories. The desired point is $x^{*}=0$ at $t^{*}=1$ with $\sigma^{*}=0.01$. The chosen parameters are $g=0.25$, $S=N$, $\alpha(t) = 0.5-0.5\tanh(10(t-0.6))$}}
    \label{fig:adapt_toy_result}
\end{figure}

\corr{
Fig. \ref{fig:adapt_toy_result} shows the adaptation of a model learnt using the KLE method. The data $\mathscr{D}$ consist of $N=10$ scalar trajectories. The original mean shown on the right of Fig. \ref{fig:adapt_toy_result} was obtained with a bandwidth parameter $h=0.03$. The new desired end point is $x^{*}=0$ at $t^{*}=1$ with $\sigma^{*}=0.01$. To adapt the model, the data set $\mathscr{E}$ is generated with $S=10N$ samples and its related kernel has bandwidth $g=0.25$. The activation function is has the form $\alpha(t)=0.5-0.5\tanh(a(t-b))$, with $a=10$ and $b=0.6$. The adapted model estimates $\hat{\mu}(t^{*}) = 0.001$ and $\hat{\sigma}(t^{*}) = 0.0182$.
}

\begin{figure}[ht]
    \centering
    \includegraphics[width=0.5\textwidth]{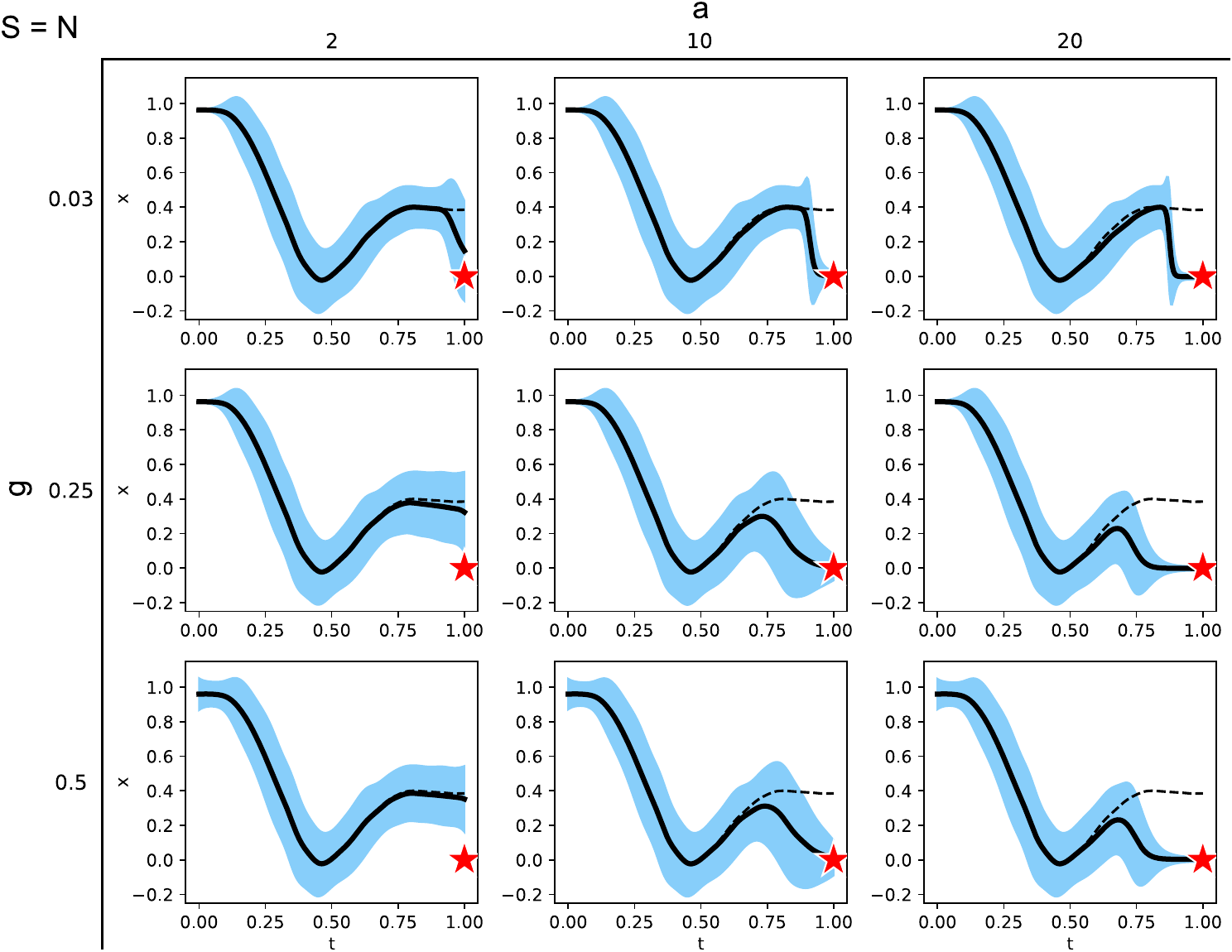}
    \caption{\corr{Effect of changing parameters $g$ and $a$ in a scalar adaptation task with $S=N$.}}
    \label{fig:adapt_toy_1}
\end{figure}

\begin{figure}[ht]
    \centering
    \includegraphics[width=0.5\textwidth]{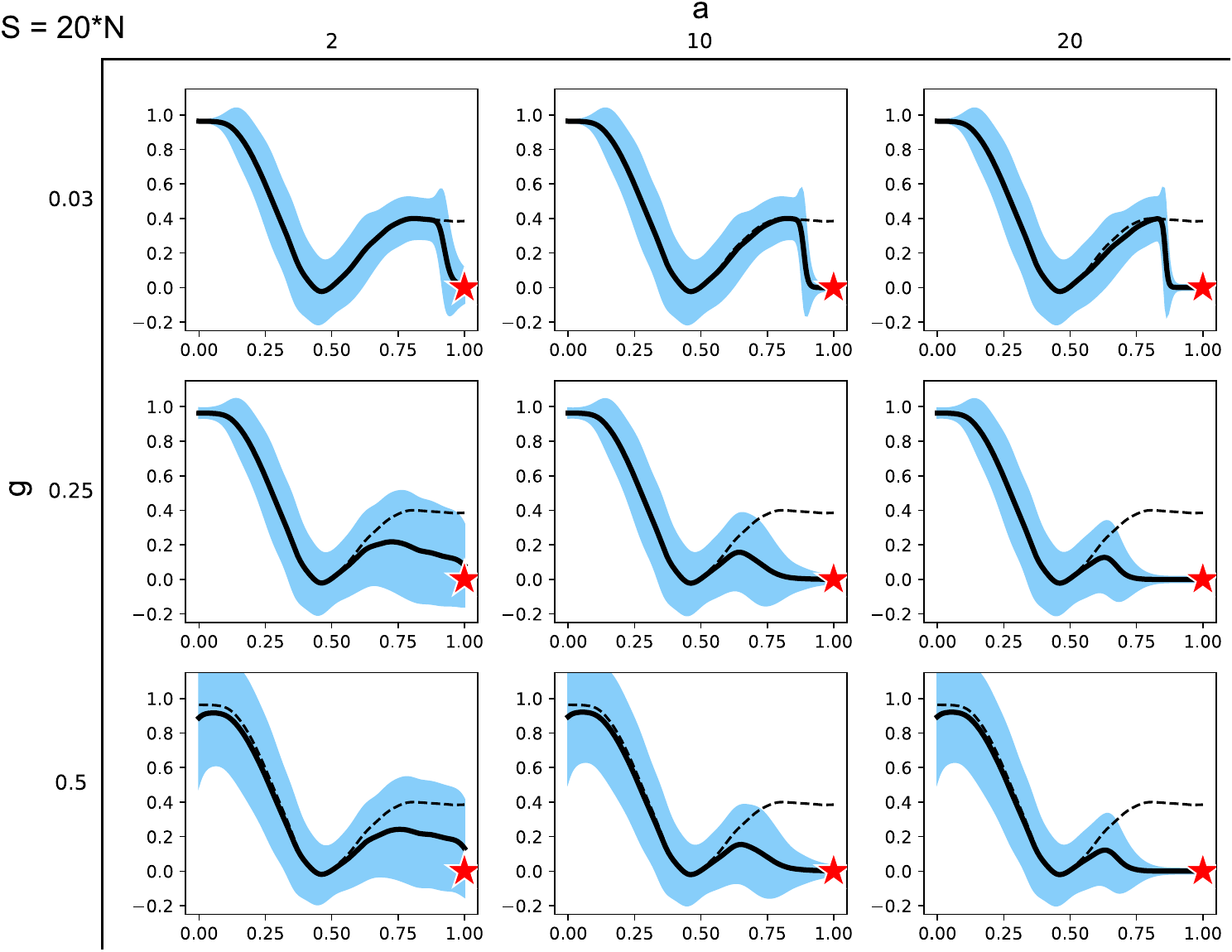}
    \caption{\corr{Effect of changing parameters $g$ and $a$ in a scalar adaptation task with $S=20N$.}}
    \label{fig:adapt_toy_2}
\end{figure}

\corr{
Figs. \ref{fig:adapt_toy_1} and \ref{fig:adapt_toy_2} show the effect of changing the values of parameters $S$, $g$, and $a$. On the other hand, $b$ changes the time of transition as it moves horizontally cut-off point of $\alpha(t)$. The effect of changing the other three parameters can be summerised as follows:
\begin{itemize}
    \item Bandwidth $g$: Large values mean the points in $\mathscr{E}$ start weighing in the regression earlier. Hence, a smoother transition is expected. However, $g$ must not be large enough to considerably affect points far away from $t^{*}$.
    \item Slope $a$: Small values give a smoother transition. However, a too small $a$ inflates the variance as the regression considers data from both data sets during the transition.
    \item Number of samples $S$: Large values produce a faster transition as the data in $\mathscr{E}$ weighs heavily in the regression. Small values produce slower transition and thus inflation of variance.
\end{itemize}
In addition, it is worth mentioning that the negative effect of variance inflation is exclusive to the case in which the adaptation point is outside the range of demonstrations.
}

\section{Eigenvalues and eigenvectors of manipulability transfer in the experiment of Sec. \ref{sec:exp_4}} \label{sec:appendix_eigen}

\corr{
The estimated $\mathbf{X}_H$ and transferred $\mathbf{X}_F$ have the following eigenvalues:
\begin{eqnarray*}
    &{}&\lambda(\mathbf{X}_H) = (1,\, 0.821,\, 0.657,\, 0.059,\, 0.035,\, 0.003),
    \\
    &{}&\lambda(\mathbf{X}_F) = (1,\, 0.722,\, 0.524,\, 0.058,\, 0.043,\, 0.007),
\end{eqnarray*}
\noindent with corresponding eigenvectors:
\[
    \begin{split}
        \mathrm{eig}(\mathbf{X}_H)&=\left\{
        \left[\begin{array}{c} -0.634 \\-0.457 \\0.530 \\-0.043 \\-0.206 \\-0.252\end{array}\right],
        \left[\begin{array}{c}-0.765\\ 0.404\\ -0.477\\ 0.039\\ 0.128\\ 0.077 \end{array}\right],
        \left[\begin{array}{c} 0.029\\ -0.688\\ -0.608\\ 0.155\\ 0.207\\ -0.298\end{array}\right],
        \right.\\&\qquad\left.
        \left[\begin{array}{c}-0.025 \\-0.080\\ 0.006\\ 0.855\\ -0.356\\ 0.367 \end{array}\right],
        \left[\begin{array}{c}-0.041\\ 0.016\\ 0.347\\ 0.345\\ 0.870\\ 0.034 \end{array}\right],
        \left[\begin{array}{c}-0.097\\ -0.384\\ -0.030\\ -0.349\\ 0.120\\ 0.840\end{array}\right]
        \right\}
    \end{split}
\]
\[
    \begin{split}
        \mathrm{eig}(\mathbf{X}_F)&=\left\{
        \left[\begin{array}{c} -0.138\\ -0.095\\ 0.936\\ -0.077\\ -0.298\\ -0.045\end{array}\right],
        \left[\begin{array}{c} 0.985\\ 0.053\\ 0.159\\ 0.019\\ 0.016\\ 0.044\end{array}\right],
        \left[\begin{array}{c} 0.072\\ -0.963\\ -0.074\\ 0.073\\ 0.058\\ -0.229\end{array}\right],
        \right.\\&\qquad\left.
        \left[\begin{array}{c} -0.041\\ -0.036\\ 0.056\\ 0.912\\ -0.088\\ 0.392\end{array}\right],
        \left[\begin{array}{c} -0.063\\ 0.077\\ 0.297\\ 0.140\\ 0.926\\ -0.158\end{array}\right],
        \left[\begin{array}{c} -0.031\\ -0.229\\ 0.049\\ -0.370\\ 0.206\\ 0.874\end{array}\right]
        \right\}
    \end{split}
\]
}

\end{document}